# Sparse quadratic classification rules via linear dimension reduction

Irina Gaynanova and Tianying Wang
Department of Statistics, Texas A&M University
3143 TAMU, College Station, TX 77843

## Abstract

We consider the problem of high-dimensional classification between the two groups with unequal covariance matrices. Rather than estimating the full quadratic discriminant rule, we propose to perform simultaneous variable selection and linear dimension reduction on original data, with the subsequent application of quadratic discriminant analysis on the reduced space. In contrast to quadratic discriminant analysis, the proposed framework doesn't require estimation of precision matrices and scales linearly with the number of measurements, making it especially attractive for the use on high-dimensional datasets. We support the methodology with theoretical guarantees on variable selection consistency, and empirical comparison with competing approaches. We apply the method to gene expression data of breast cancer patients, and confirm the crucial importance of ESR1 gene in differentiating estrogen receptor status.

## 1 Introduction

We consider a binary classification problem: given $n$ independent pairs $(X_i, Y_i)$ from a joint distribution $(X, Y)$ on $\mathbb{R}^p \times \{1, 2\}$, our goal is to both learn a rule that will assign one of two labels to a new data point $X \in \mathbb{R}^p$, and determine the subset of $p$ variables that influences the rule. One of the popular classification tools is linear discriminant analysis, or LDA (Mardia et al., 1979, Chapter 11). While it gives unsatisfactory results when applied to high-dimensional datasets (Dudoit et al., 2002), recent work suggests that additional regularization, variable selection in particular, leads to dramatic performance improvements. Earlier approaches perform variable selection and regularize the sample covariance matrix by treating it as diagonal (Tibshirani et al., 2003; Witten and Tibshirani, 2011). More recent methods directly estimate the discriminant directions by using convex optimization framework with sparsity-inducing penalties (Cai and Liu, 2011; Mai et al., 2012; Gaynanova et al., 2016).

Despite these significant advances, a key underlying assumption of linear discriminant analysis is the equality of covariance matrices between the groups, $\Sigma_1 = \Sigma_2$. This assumption is unlikely to be satisfied in practice, leading to suboptimal performance of linear rule.

When the measurements are normally distributed, $X_i|Y_i = g \sim \mathcal{N}(\mu_g, \Sigma_g)$, $g \in \{1, 2\}$, with $\Sigma_1 \neq \Sigma_2$, the Bayes rule is quadratic, leading to quadratic discriminant analysis, or QDA. As with linear case, the quadratic discriminant analysis performs poorly when $p$ is large. This unsatisfactory performance is largely due to the estimation of precision matrices $\Sigma_1^{-1}$ and $\Sigma_2^{-1}$, a task that is extremely challenging when $p \gg n$. In fact, even when $p = n/2$ and the assumption of equal covariance matrices is violated, the misclassification error rate of sample QDA is worse than the rates of regularized linear discriminant methods (Gaynanova et al., 2016, supplement).

Several extensions of sample QDA have been proposed. A common strategy is to jointly estimate $\Sigma_1^{-1}$ and $\Sigma_2^{-1}$. Friedman (1989); Ramey et al. (2016) regularize sample covariance matrices by shrinkage. Wu et al. (2018) impose equicorrelation structure on each covariance matrix by pooling both the diagonal and off-diagonal elements. Danaher et al. (2014); Guo et al. (2011); Price et al. (2014); Simon and Tibshirani (2011) use a penalized likelihood technique, where the penalty enforces similarity either between the covariance matrices $\Sigma_g$ or the precision matrices $\Sigma_g^{-1}$. While these methods perform better than quadratic rules based on sample covariance matrices, they again rely on estimating two precision matrices. As such, additional assumptions on $\Sigma_g^{-1}$ such as sparsity are usually enforced, and the estimation procedure scales quadratically with the number of measurements $p$. Moreover, the resulting classification rules still rely on all $p$ variables, and therefore can not be used for both classification and variable selection.

Li and Shao (2015) address the variable selection problem by enforcing sparsity in both the covariance matrices and the vector of mean differences via thresholding. The method comes with strong theoretical guarantees on classification consistency and promising empirical performance. Nevertheless, it again requires additional assumptions on $\Sigma_g$, and is computationally prohibitive for large $p$ due to required matrices inversion together with a 3-dimensional search over tuning parameter values.

In summary, a significant progress in linear discriminant methods made it possible to apply them to large datasets and perform variable selection. In practice, however, the covariance matrices are often unequal, but the existing quadratic methods typically can not perform variable selection, and are computationally prohibitive for large $p$. In this work we bridge the gap between the linear and the quadratic methods by developing a new classification rule that takes into account unequal covariance matrices without sacrificing either variable selection or the computational speed.

Our key methodological contribution is a different approach for constructing quadratic rule in high-dimensional settings compared to the ones taken in the literature. The existing methods rely on improved estimation of the full Bayes quadratic discriminant rule by exploring additional structural assumptions on $\Sigma_g$ or $\Sigma_g^{-1}$ (Simon and Tibshirani, 2011; Price et al., 2014; Le and Hastie, 2014; Li and Shao, 2015; Wu et al., 2018). In contrast, we modify the Fisher's formulation of linear discriminant analysis for the case of unequal covariance matrices. The resulting method performs simultaneous variable selection and projection of original data on a lower-dimensional space, with the subsequent application of quadratic discriminant analysis. We call this approach discriminant analysis via projections, or DAP.

Unlike the existent quadratic methods, our rule is linear in $p$, which allows us to devise a very efficient optimization procedure to simultaneously estimate the projection directions and perform variable selection. For $p = 500$, it takes around 1.5 seconds to implement our method, whereas the closest competing sparse quadratic method takes 30 minutes. This makes it possible to apply our approach in situations where other quadratic methods are computationally infeasible. Moreover, we connect the variables in our rule with the nonzero variables in the linear part of Bayes quadratic rule, and prove the variable selection consistency of our method in high-dimensional settings. Empirical studies confirm that for large values of $p$ the proposed rule leads to competitive, and often smaller, misclassification error rates than the existing approaches. At the same time, our method consistently selects the sparsest models thus achieving the best balance between model complexity and misclassification error rate. Finally, the application to gene expression data of breast cancer patients (Chin et al., 2006) confirms the crucial importance of ESR1 gene in differentiating estrogen receptor status; an insight that is not possible with other approaches due to much higher complexity of corresponding classification rules.

The rest of this paper is organized as follows. In Section 2, we describe a new quadratic classification rule, discriminant analysis via projections. We connect the proposed approach to both linear and quadratic discriminant analysis, and derive an efficient optimization algorithm for sparse estimation. In Section 3, we provide theoretical guarantees on the variable selection consistency of our method in high-dimensional settings. In Section 4, we conduct empirical studies on both simulated and real data. In Section 5, we discuss possible extensions in future work.

**Notation:** For a vector $v \in \mathbb{R}^p$, we let $\|v\|_1 = \sum_{i=1}^p |v_i|$, $\|v\|_2 = (\sum_{i=1}^p v_i^2)^{1/2}$, $\|v\|_\infty = \max_i |v_i|$. We use $e_j$ to denote a unit norm vector with $j$th element being equal to one, and $1_p$ to denote the vector of ones of length $p$. For a matrix $M \in \mathbb{R}^{n \times p}$, we let $\|M\|_{\infty,2} = \max_{1 \le i \le n}(\sum_{j=1}^p m_{ij}^2)^{1/2}$, $\|M\|_2 = \sup_{x:\|x\|_2=1} \|Mx\|_2$ and $|M|$ be the determinant of $M$. Given an index set $A$, we use $M_A$ to denote the submatrix of $M$ with columns indexed by $A$. For a square matrix $M$, we use $M_{AA}$ to denote the submatrix of $M$ with both rows and columns indexed by $A$. We use $I$ to denote the identity matrix. We use $a_n \lesssim b_n$ to denote that there exists a constant $C > 0$ such that $a_n \le Cb_n$ for $n$ sufficiently large. We also let $a \vee b = \max(a, b)$.

# 2 Discriminant analysis via projections

## 2.1 Review of Fisher's discriminant analysis

Consider $n$ independent pairs $(X_i, Y_i)$ from a joint distribution $(X, Y)$ on $\mathbb{R}^p \times \{1, 2\}$. Let $\Sigma_g = \text{cov}(X|Y = g)$, $g = 1, 2$ and assume the covariance matrices are equal, $\Sigma_1 = \Sigma_2$. Fisher's discriminant analysis seeks a linear combination of $p$ measurements that maximize between group variability with respect to within group variability (Mardia et al., 1979, Chapter 11):

$$\underset{v \in \mathbb{R}^p}{\text{maximize}} \left\{ \frac{v^{\text{T}}(\bar{x}_1 - \bar{x}_2)(\bar{x}_1 - \bar{x}_2)^{\text{T}} v}{v^{\text{T}} W v} \right\}, \tag{1}$$

where $W = (n - 2)^{-1} \sum_{g=1}^{2} (n_g - 1) S_g$ is the pooled sample covariance matrix, $S_g$ is the sample covariance matrix for group $g$, $n_g$ is the number of samples in group $g$, and $\bar{x}_g$ is the sample mean for group $g$. Letting $\widehat{v}$ be a vector at which the maximum above is achieved, the resulting classification rule for a new observation with observed value $x \in \mathbb{R}^p$ is

$$h_{\widehat{v}}(x) = \underset{g \in \{1,2\}}{\text{argmin}} \left\{ (x^{\text{T}}\widehat{v} - \bar{x}_g^{\text{T}}\widehat{v})^{\text{T}} (\widehat{v}^{\text{T}} W \widehat{v})^{-1} (x^{\text{T}}\widehat{v} - \bar{x}_g^{\text{T}}\widehat{v}) - 2 \log(n_g/n) \right\}. \tag{2}$$

Hence, both the new observation $x \in \mathbb{R}^p$ and the data $X \in \mathbb{R}^{n \times p}$ are projected onto the line determined by $\widehat{v}$, and the classification is performed according to Mahalanobis distance to the class means in the projected space. Since both the objective function in (1) and the classification rule (2) are invariant to the scaling of discriminant vector $\widehat{v}$, it can be expressed as $\widehat{v} = cW^{-1}(\bar{x}_1 - \bar{x}_2)$ for any constant $c \neq 0$. Moreover, the Fisher's rule (2) coincides with sample plug-in Bayes rule under the normality assumption, that is $X_i|Y_i = g \sim N(\mu_g, \Sigma)$.

## 2.2 Modification of Fisher's rule for the case of unequal covariance matrices

Our proposal is based on the modification of criterion (1) to the case of unequal covariance matrices. Specifically, we consider two discriminant directions instead of one

$$\widehat{v}_g = \underset{v_g \in \mathbb{R}^p}{\text{argmax}} \left\{ \frac{v_g^{\text{T}}(\bar{x}_1 - \bar{x}_2)(\bar{x}_1 - \bar{x}_2)^{\text{T}} v_g}{v_g^{\text{T}} S_g v_g} \right\} \quad (g = 1, 2). \tag{3}$$

Similar to Fisher's criterion, the solutions to (3) can be expressed as $\widehat{v}_g = c_g S_g^{-1}(\bar{x}_1 - \bar{x}_2)$ for any $c_g \neq 0$, $g = 1, 2$. Subsequently, given matrix $\widehat{V} = [\widehat{v}_1 \ \widehat{v}_2]$, we modify rule (2) to take into account unequal covariance matrices as

$$h_{\widehat{V}}(x) = \underset{g \in \{1,2\}}{\text{argmin}} \left\{ (x - \bar{x}_g)^{\text{T}} \widehat{V} (\widehat{V}^{\text{T}} S_g \widehat{V})^{-1} \widehat{V}^{\text{T}} (x - \bar{x}_g) + \log |\widehat{V}^{\top} S_g \widehat{V}| - 2 \log(n_g/n) \right\}. \tag{4}$$

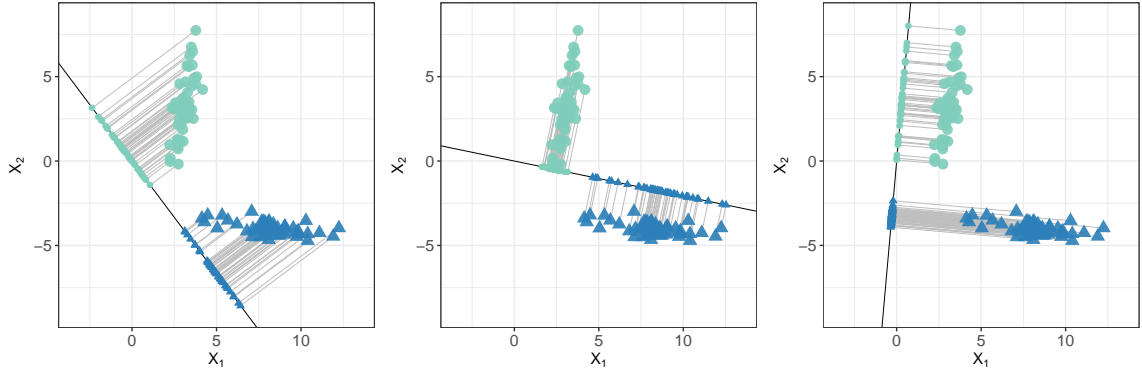

Figure 1: Two-group classification problem with $p = 2$ and unequal covariance matrices. **Left:** Projection using Fisher's discriminant vector. **Middle:** Projection using the covariance structure from the 1st group (circles). **Right:** Projection using the covariance structure from the 2nd group (triangles).

**Remark 1.** *If $\widehat{v}_1$ and $\widehat{v}_2$ are linearly dependent, then $\widehat{V}$ has rank one, and $\widehat{V}^{\mathrm{T}} S_1 \widehat{V}$ and $\widehat{V}^{\mathrm{T}} S_2 \widehat{V}$ are both singular. In this case the subspace spanned by the columns of $\widehat{V}$ is the same as the subspace spanned by only one column, and we use $\widehat{V} = \widehat{v}_1$ in (4).*

Rule (4) is equivalent to applying quadratic discriminant rule to $\widehat{V}^{\mathrm{T}} x$ instead of applying it directly to $x$. Unlike the equivalence between the Fisher's rule and the linear discriminant rule, in Section 2.6 we show that rule (4) is generally not equivalent to quadratic discriminant analysis. Nevertheless, formulation (4) allows to overcome possible rank degeneracy of $S_g$ as well as perform variable selection. First, rule (4) requires inversion of $2 \times 2$ matrices $\widehat{V}^{\mathrm{T}} S_g \widehat{V}$, which are likely to be positive definite, in contrast to $S_g$. Secondly, since (4) effectively applies quadratic rule to $\widehat{V}^{\mathrm{T}} x$ instead of $x$, it only relies on those variables for which the corresponding rows of $\widehat{V}$ are nonzero. Hence, performing variable selection is equivalent to using row-sparse matrix $\widehat{V}$. Figure 1 shows that each $\widehat{v}_g$ from (3) can be viewed as a basis vector for the reduced space, and coincides with discriminant vector $\widehat{v}$ in Fisher's rule (1) if the pooled sample covariance matrix $W = S_1 = S_2$. Therefore, we call rule (4) the discriminant analysis via projections.

## 2.3 Sparse estimation

While rule (4) allows to overcome the potential singularity of sample covariance matrices, it still requires estimation of $\mathcal{O}(p)$ parameters and therefore may lead to poor performance in the high-dimensional settings when $p \gg n$. At the same time, in the context of linear discriminant analysis the classification performance can be significantly improved by directly estimating the discriminant vector with sparsity regularization (Cai and Liu, 2011; Mai et al., 2012). Guided by this intuition, our goal is to obtain sparse estimates of $\psi_1 = c_1 \Sigma_1^{-1} \delta$ and $\psi_2 = c_2 \Sigma_2^{-1} \delta$ with $\delta = \mu_1 - \mu_2$, which are the population counterparts of $\widehat{v}_1$ and $\widehat{v}_2$ in (3). This approach leads to regularized row-sparse $\widehat{V}$ that can be used directly in rule (4).

To produce sparse estimates of $\psi_1$ and $\psi_2$, we consider penalized empirical risk minimization framework:

$$\widehat{V} = [\widehat{v}_1\ \widehat{v}_2] = \underset{v_1, v_2 \in \mathbb{R}^p}{\operatorname{argmin}} \left\{ \widehat{L}_{\psi_1}(v_1) + \widehat{L}_{\psi_2}(v_2) + \lambda \operatorname{Pen}(V) \right\},$$

where $\widehat{L}_{\psi_1}(v_1)$, $\widehat{L}_{\psi_2}(v_2)$ are empirical loss functions associated with $\psi_1$, $\psi_2$; $\lambda > 0$ is the tuning parameter, and $\operatorname{Pen}(V)$ is the sparsity-inducing penalty.

**Remark 2.** *Another possibility is to add sparse penalization directly within criterion (3). In linear discriminant analysis, this approach leads to significant improvement over sample plug-in rule (Witten and Tibshirani, 2011). However, it also leads to nonconvex optimization problem and potential difficulties in obtaining very sparse solutions (Gaynanova et al., 2017). Therefore, we do not pursue the direct penalization here.*

First, we discuss our choice of penalty. As we are interested in simultaneous variable selection, that is row-sparsity of $\widehat{V}$, we propose to use group penalty. Specifically, we choose group-lasso, $\operatorname{Pen}(V) = \sum_{j=1}^{p}(v_{1j}^2 + v_{2j}^2)^{1/2}$ (Yuan and Lin, 2006), due to its convexity. Other possibilities include nonconvex group penalties, we refer the reader to Huang et al. (2012) for the review.

Next, we discuss our choice of empirical loss functions $\widehat{L}_{\psi_1}(v_1)$ and $\widehat{L}_{\psi_2}(v_2)$. Both the criterion (3) and the rule (4) are invariant to the scale of $\widehat{V}$, that is to the choice of constants $c_1$ and $c_2$. While the naive approach is to fix $c_1 = c_2 = 1$, we use $c_1 = \pi_2/(1+\pi_2^2 \delta^T \Sigma_1^{-1} \delta)$, $c_2 = \pi_1/(1+\pi_1^2 \delta^T \Sigma_2^{-1} \delta)$, which lead to lower-bounded empirical loss function as well as significant computational savings. To be specific, we take advantage of the following equivalence due to the Sherman–Morrison formula:

**Proposition 1.** *For any $\rho \neq 0$, any non-singular matrix $M \in \mathbb{R}^{p \times p}$ and any vector $a \in \mathbb{R}^p$*

$$(M + \rho^2 aa^T)^{-1} \rho a = \rho M^{-1} a (1 + \rho^2 a^T M^{-1} a)^{-1} \propto M^{-1} a.$$

Our choice of $c_1$ and $c_2$ leads to $\psi_1 = (\Sigma_1 + \pi_2^2 \delta \delta^T)^{-1} \pi_2 \delta$ and $\psi_2 = (\Sigma_2 + \pi_1^2 \delta \delta^T)^{-1} \pi_1 \delta$. Consider the following quadratic loss function associated with $\psi_1$

$$L_{\psi_1}(v_1) = (v_1 - \psi_1)^T (\Sigma_1 + \pi_2^2 \delta \delta^T)(v_1 - \psi_1)/2 = v_1^T \Sigma_1 v_1/2 + (\pi_2 \delta^T v_1 - 1)^2/2 + C,$$

where $C$ is a constant independent of $v_1$. Consider the empirical version of this loss function

$$\widehat{L}_{\psi_1}(v_1) = v_1^T S_1 v_1/2 + \left(n^{-1} n_2 d^T v_1 - 1\right)^2/2 + C, \tag{5}$$

where $d = \bar{x}_1 - \bar{x}_2$. First, $\widehat{L}_{\psi_1}(v_1)$ is invariant under linear transformation of the data (Rukhin, 1992). Secondly, $\widehat{L}_{\psi_1}(v_1)$ is always bounded from below by $C$, even when $S_1$ is singular. This ensures guaranteed convergence of the block-coordinate descent algorithm without the need to regularize $S_1$, and in particular, is not the case for $c_1 = 1$.

Furthermore, let $X_1 \in \mathbb{R}^{n_1 \times p}$ be the submatrix of $X$ corresponding to the 1st group, and $X_2 \in \mathbb{R}^{n_2 \times p}$ be the one corresponding to the 2nd group. Let $X$ be column-centered so that $\bar{x} = n^{-1}(n_1\bar{x}_1 + n_2\bar{x}_2) = 0$, and hence $d = n_2^{-1} n \bar{x}_1$. Then the loss (5) can be rewritten as

$$\widehat{L}_{\psi_1}(v_1) = v_1^\mathrm{T} S_1 v_1/2 + (\bar{x}_1^\mathrm{T} v_1 - 1)^2/2 + C = n_1^{-1} v_1^\mathrm{T} X_1^\mathrm{T} X_1 v_1/2 - v_1^\mathrm{T} \bar{x}_1 + C$$
$$= n_1^{-1} \|X_1 v_1 - 1_{n_1}\|_2^2/2 + C.$$

That is, the loss function can be expressed as the linear regression loss function. Similarly,

$$\widehat{L}_{\psi_2}(v_2) = n_2^{-1} \|X_2 v_2 + 1_{n_2}\|_2^2/2 + C.$$

Therefore, our choice of $c_1$ and $c_2$ allows to re-express the problem of estimating $\psi_1$ and $\psi_2$ as a regression problem. This leads to efficient optimization algorithm described in Section 2.4.

In summary, given the column-centered data matrix $X \in \mathbb{R}^{n \times p}$ with submatrices $X_1 \in \mathbb{R}^{n_1 \times p}$, $X_2 \in \mathbb{R}^{n_2 \times p}$ corresponding to two groups, we find $\widehat{V} = [\widehat{v}_1\ \widehat{v}_2] \in \mathbb{R}^{p \times 2}$ as the solution to

$$\underset{V=[v_1,v_2]\in\mathbb{R}^{p\times 2}}{\text{minimize}} \left\{ n_1^{-1}\|X_1 v_1 - 1_{n_1}\|_2^2/2 + n_2^{-1}\|X_2 v_2 + 1_{n_2}\|_2^2/2 + \lambda \sum_{j=1}^{p} (v_{1j}^2 + v_{2j}^2)^{1/2} \right\}. \quad (6)$$

If $\lambda = 0$, $\widehat{V}$ coincides with the solution to (3) up to the choice of scaling. If $\lambda > 0$, then $\widehat{V}$ is row-sparse leading to variable selection. Given $\widehat{V}$, we apply rule (4) for classification.

## 2.4 Optimization algorithm

In this section we derive a block-coordinate descent algorithm to solve (6). Consider the optimality conditions with respect to each block $v_j = (v_{1j}, v_{2j})^\mathrm{T}$ (Boyd and Vandenberghe, 2004, Chapter 5):

$$n_1^{-1} X_{1j}^\mathrm{T} X_{1j} v_{1j} = n_1^{-1} X_{1j}^\mathrm{T} (1_{n_1} - \sum_{k \neq j} v_{1k} X_{1k}) - \lambda u_{1j},$$
$$n_2^{-1} X_{2j}^\mathrm{T} X_{2j} v_{2j} = n_2^{-1} X_{2j}^\mathrm{T} (-1_{n_2} - \sum_{k \neq j} v_{2k} X_{2k}) - \lambda u_{2j};$$

where $u_j = (u_{1j}, u_{2j})^\mathrm{T}$ is the subgradient of $(v_{1j}^2 + v_{2j}^2)^{1/2}$

$$u_j = \begin{cases} v_j/\|v_j\|_2, & \text{if } \|v_j\|_2 \neq 0; \\ \in \{u : \|u\|_2 \leq 1\}, & \text{if } \|v_j\|_2 = 0. \end{cases} \quad (7)$$

In general, $n_1^{-1} X_{1j}^\mathrm{T} X_{1j} \neq n_2^{-1} X_{2j}^\mathrm{T} X_{2j}$, hence the block-update is not available in closed form and requires a line search (Barber and Drton, 2010). However, guided by the computational considerations as well as the ideas of standardized group lasso (Simon and Tibshirani, 2012), we pre-standardize $X_1$ and $X_2$ so that $n_1^{-1}\mathrm{diag}(X_1^\mathrm{T} X_1) = n_2^{-1}\mathrm{diag}(X_2^\mathrm{T} X_2) = 1_p$, and then perform the back-scaling of $\widehat{v}_1$, $\widehat{v}_2$. This ensures that the penalization of different variables is independent of their relative scales. Finally, we are ready to present the algorithm.

Define the residual vectors $r_1$, $r_2$ as

$$r_{1j} = n_1^{-1} X_{1j}^\mathrm{T} (1_{n_1} - \sum_{l=1}^p v_{1l} X_{1l}), \quad r_{2j} = n_2^{-1} X_{2j}^\mathrm{T} (-1_{n_2} - \sum_{l=1}^p v_{2l} X_{2l});$$

with $r_j = (r_{1j}, r_{2j})^\mathrm{T}$. From the optimality conditions, the equations for the $j$th block $v_j = (v_{1j}, v_{2j})^\mathrm{T}$ can be rewritten as

$$v_j = (1 - \lambda/\|v_j + r_j\|_2)_+ (v_j + r_j),$$

where $a_+ = \max(0, a)$. Starting with some initial value $V^{(0)}$, the block-coordinate descent algorithm proceeds by iterating the updates of $v_1$, $v_2$ with updates of residuals $r_1, r_2$ until convergence. Due to convexity of (6), the boundedness of the objective function from below, and the separability of the penalty with respect to block updates, the global optimum is finite and the algorithm is guaranteed to converge to the global optimum from any starting point (Tseng, 2001).

## 2.5 Connection with sparse linear discriminant analysis

We show that the sparse linear discriminant analysis can be viewed as a very special case of the proposed approach.

**Proposition 2.** *Consider the sparse discriminant analysis in Gaynanova et al. (2016) that finds the discriminant vector $\widetilde{v}(\lambda)$ for a given value of tuning parameter $\lambda > 0$. Define $c = (n_1/n)^{1/2} + (n_2/n)^{1/2}$. Under the constraint $(n/n_1)^{1/2} v_1 = (n/n_2)^{1/2} v_2$, the solution to (6) satisfies*

$$(n/n_1)^{1/2} \widehat{v}_1(\lambda) = (n/n_2)^{1/2} \widehat{v}_2(\lambda) = c\,\widetilde{v}(\lambda/c).$$

When $v_1$ and $v_2$ are restricted to be in the same direction, (6) gives the same solution as the sparse linear discriminant analysis up to scaling.

## 2.6 Connection with quadratic discriminant analysis

Let $Y$ be a group indicator such that $P(Y = 1) = \pi_1$ and $P(Y = 2) = 1 - \pi_1 = \pi_2$, and consider $X|Y = g \sim N(\mu_g, \Sigma_g)$ ($g = 1, 2$). The Bayes rule assigns a new observation with observed value $x \in \mathbb{R}^p$ to group one if and only if

$$\begin{aligned} & x^\mathrm{T}(\Sigma_2^{-1} - \Sigma_1^{-1})x - 2x^\mathrm{T}(\Sigma_2^{-1}\mu_2 - \Sigma_1^{-1}\mu_1) \\ & \quad + \log\left(|\Sigma_2|/|\Sigma_1|\right) - \mu_1^\mathrm{T}\Sigma_1^{-1}\mu_1 + \mu_2^\mathrm{T}\Sigma_2^{-1}\mu_2 + 2\log(\pi_1/\pi_2) > 0. \end{aligned} \quad (8)$$

Consider centering $x$ by the overall mean $\mathbb{E}(X) = \mu = \pi_1\mu_1 + \pi_2\mu_2$.

**Proposition 3.** *Let $\delta = \mu_1 - \mu_2$. The Bayes rule (8) can be written as*

$$\begin{aligned}
&(x-\mu)^{\mathrm{T}}(\Sigma_2^{-1} - \Sigma_1^{-1})(x-\mu) + \log\left(|\Sigma_2|/|\Sigma_1|\right) \\
&+ 2(x-\mu)^{\mathrm{T}}(\pi_1\Sigma_2^{-1}\delta + \pi_2\Sigma_1^{-1}\delta) + \pi_1^2\delta^{\mathrm{T}}\Sigma_2^{-1}\delta - \pi_2^2\delta^{\mathrm{T}}\Sigma_1^{-1}\delta + 2\log(\pi_1/\pi_2) > 0.
\end{aligned} \quad (9)$$

Consider the population version of the proposed discriminant analysis via projections, that is applying Bayes rule to $\Psi^{\mathrm{T}}X$ with $\Psi^{\mathrm{T}}X|Y = g \sim N(\Psi^{\mathrm{T}}\mu_g, \Psi^{\mathrm{T}}\Sigma_g\Psi)$ and $\Psi = [\psi_1, \psi_2] = [c_1\Sigma_1^{-1}\delta, \ c_2\Sigma_2^{-1}\delta]$, $c_1, c_2 \neq 0$.

**Proposition 4.** *Consider the population version of rule (4), that is substituting $\Psi$ for $\widehat{V}$, $\Sigma_g$ for $S_g$, $\mu_g$ for $\bar{x}_g$ and $\pi_g$ for $n_g/n$. A new observation with value $x$ is assigned to group one if and only if*

$$\begin{aligned}
&(x-\mu)^{\mathrm{T}}\Psi\left\{(\Psi^{\mathrm{T}}\Sigma_2\Psi)^{-1} - (\Psi^{\mathrm{T}}\Sigma_1\Psi)^{-1}\right\}\Psi^{\mathrm{T}}(x-\mu) + \log\left(|\Psi^{\mathrm{T}}\Sigma_2\Psi|/|\Psi^{\mathrm{T}}\Sigma_1\Psi|\right) \\
&+ 2(x-\mu)^{\mathrm{T}}(\pi_1\Sigma_2^{-1}\delta + \pi_2\Sigma_1^{-1}\delta) + \pi_1^2\delta^{\mathrm{T}}\Sigma_2^{-1}\delta - \pi_2^2\delta^{\mathrm{T}}\Sigma_1^{-1}\delta + 2\log(\pi_1/\pi_2) > 0.
\end{aligned} \quad (10)$$

The only difference between the rules in Proposition 3 and 4 is on the first line, which involves the quadratic and the log terms. The linear terms and the remaining constant terms are identical. Therefore, rule (10) can be viewed as an approximation to rule (9).

While rule (10) is not the same as the Bayes rule, and therefore will lead to inferior performance on the population level, in Section 4 we see this relationship to be reversed when the corresponding regularized sample versions are considered and $p$ is large relative to the sample size $n$. The main advantage of rule (10) comes from the significant reduction in the number of parameters to be estimated. Specifically, matrix $\Psi$ has $p \times 2$ elements leading to $\mathcal{O}(p)$ parameters in rule (10). In contrast, the Bayes rule requires estimation of the $\Sigma_2^{-1} - \Sigma_1^{-1}$ leading to $\mathcal{O}(p^2)$ parameters in total.

# 3 Variable selection consistency in high-dimensional settings

We establish the variable selection consistency of estimator in (6) under the following assumptions.

**Assumption 1** (Normality). $X_i|Y_i = g \sim \mathcal{N}(\mu_g, \Sigma_g)$, $\mathrm{pr}(Y_i = g) = \pi_g$ *for $g = 1, 2$ with $0 < \pi_{\min} \leq \pi_1/\pi_2 \leq \pi_{\max} < 1$.*

**Assumption 2** (Sparsity). *Let $\delta = \mu_1 - \mu_2$, $A = \{i : (e_i^{\mathrm{T}}\Sigma_1^{-1}\delta)^2 + (e_i^{\mathrm{T}}\Sigma_2^{-1}\delta)^2 \neq 0\}$, $A^c = \{1,\ldots,p\}/A$ and $\mathrm{card}(A) = s$. That is, $A$ is the index set of nonzero variables in $\Sigma_1^{-1}\delta$ or $\Sigma_2^{-1}\delta$.*

**Assumption 3** (Irrepresentability). *There exist $\alpha \in (0, 1]$ such that*

$$\max_{\substack{u_1, u_2 \in \mathbb{R}^s \\ u_{1i}^2 + u_{2i}^2 \leq 1 \; \forall i}} \|\Sigma_{1A^cA}\Sigma_{1AA}^{-1}u_1, \Sigma_{2A^cA}\Sigma_{2AA}^{-1}u_2\|_{\infty,2} \leq 1 - \alpha.$$

**Assumption 4.** $0 < c \leq \lambda_{\min}(\Sigma_{gAA}) \leq \lambda_{\max}(\Sigma_{gAA}) \leq C$ and $e_j^{\mathrm{T}}\Sigma_g e_j \leq M$ for $j = 1, \ldots, p$ and $g = 1, 2$.

Assumption 1 is a standard assumption in the context of discriminant analysis (Mai et al., 2012; Kolar and Liu, 2015; Gaynanova and Kolar, 2015), and Assumptions 2–3 are typical in establishing variable selection consistency of penalized estimators in high-dimensional settings (Bach, 2008; Wainwright, 2009; Obozinski et al., 2011). We use Assumption 4 for convenience of treating the parameters depending on $\Sigma_g$ as constants and presenting the rates in Theorems 1 and 2 through only $n$, $p$ and $s$. We refer the reader to the Supplementary Material for the more general statements of Theorems 1 and 2 without the use of Assumption 4. To prove variable selection consistency of estimator in (6), we use the primal-dual witness technique (Wainwright, 2009). First, we prove that under the appropriate scaling of the sample sizes, and sufficiently large value of the tuning parameter $\lambda$, the variables in $A^c$ are set to zero with high probability. Let $\widehat{A} = \{i : \widehat{v}_{1i}^2 + \widehat{v}_{2i}^2 \neq 0\}$ denote the support of the solution to (6).

**Theorem 1.** *Let Assumptions 1–4 hold, the sample sizes satisfy $\min_g n_g \gtrsim s \log\{(p-s)\eta^{-1}\}$ for some $\eta \in (0, 1)$, and the tuning parameter satisfy $\lambda \gtrsim [\log\{(p-s)\eta^{-1}\}/n]^{1/2}$. Then $\mathrm{pr}(\widehat{A} \subseteq A) \geq 1 - \eta$.*

Next, we show that under the additional assumption on the minimal signal strength defined as

$$\psi_{\min} = \min_{j \in A} \left\{\pi_2^2(e_j^{\mathrm{T}}\Sigma_1^{-1}\delta)^2 + \pi_1^2(e_j^{\mathrm{T}}\Sigma_2^{-1}\delta)^2\right\}^{1/2},$$

the true variables are nonzero with high probability leading to perfect recovery. In sparse linear models this assumption is often called $\beta$-min condition (Wainwright, 2009). According to Proposition 3, $\psi_{\min}$ can be interpreted as the smallest magnitude of the nonzero variables in the linear part of the Bayes quadratic discriminant rule.

**Theorem 2.** *Let the conditions of Theorem 1 hold and $\psi_{\min} \gtrsim \lambda s^{1/2}(\max_g \delta_A^{\mathrm{T}}\Sigma_{gAA}^{-1}\delta_A \vee 1)$. Then $\mathrm{pr}(\widehat{A} = A) \geq 1 - \eta$.*

Theorem 2 reveals the advantage of using the group penalty in joint sparse estimation of $\psi_1$ and $\psi_2$. If variable $j$ is nonzero in both $\psi_1$ and $\psi_2$, then it is sufficient to have large signal in only one of $\psi_g$ for minimal signal strength condition to hold. In contrast, separate estimation via the lasso penalty will lead to the requirement of sufficiently large signal in both $\psi_1$ and $\psi_2$ simultaneously.

# 4 Empirical studies

## 4.1 Simulated data

We compare the misclassification error rates and variable selection performance of the following methods: (i) Sample QDA, rule (8) with plug-in estimates $\bar{x}_1, \bar{x}_2, S_1, S_2$; (ii) Sparse QDA of Le and Hastie (2014); (iii) Sparse QDA of Li and Shao (2015); (iv) Sparse QDA via ridge fusion (Price et al., 2014); (v) Logistic regression with pairwise interactions and lasso penalty on the vector of coefficients; (vi) Regularized discriminant analysis (Friedman, 1989); (vii) Sparse LDA (Mai et al., 2012; Gaynanova et al., 2016); (viii) Discriminant analysis via projections proposed in this paper, that is rule (4) with estimator from (6). The details of all methods' implementation together with tuning parameter selection criteria are described in Supplementary Materials.

We fix the sample sizes $n_1 = n_2 = 100$, the dimension $p \in \{100, 500\}$, and the group means $\mu_1 = 0_p$ and $\mu_2 = (1_5, -1_5, 0_{p-10})$. We consider the following types of covariance structures:

1. Block-equicorrelation with block size $b \in \{10, 100\}$ and $\rho \in [0, 1]$:

$$\Sigma_g = \begin{pmatrix} \rho I_b + (1-\rho) 1_b 1_b^{\mathrm{T}} & 0 \\ 0 & I_{p-b} \end{pmatrix}.$$

2. Block-autocorrelation with block size $b \in \{10, 100\}$ and $\rho \in [0, 1]$:

$$\Sigma_g = \{\Sigma_g\}_{i,j}, \quad \{\Sigma_g\}_{i,j} = \begin{cases} \rho^{|i-j|}, & (1 \le i, j \le b); \\ \mathbb{1}\{i = j\}, & (\text{otherwise}). \end{cases}$$

3. Spiked with parameters $q_1, q_2 \in \mathbb{R}^p$: $\Sigma_g = 30 q_1 q_1^{\mathrm{T}} + 2 q_2 q_2^{\mathrm{T}} + I$.

    (a) Block size $b = 10$: $q_1 = (1_5/\sqrt{5}, 0_{p-5})$, $q_2 = (0_{p-5}, 1_5/\sqrt{5}, 0_{p-10})$.
    
    (b) Block size $b = 100$: $q_1 = (1, \ldots, 100, 0_{p-100})^{\mathrm{T}}$ normalized so that $q_1^{\mathrm{T}} q_1 = 1$; $q_2 = (I - q_1 q_1^{\mathrm{T}})(100, \ldots, 1, 0_{p-100})^{\mathrm{T}}$ normalized so that $q_2^{\mathrm{T}} q_2 = 1$.

These structures are commonly used to assess the performance of discriminant analysis methods (Mai et al., 2012; Le and Hastie, 2014; Ramey et al., 2016). We use 8 combinations as described in Table 4.1, and fix the block sizes to make the Bayes error rate independent of $p$.

As expected, the sample QDA performs the worst, with misclassification error rates being larger than 40% consistently across all replications and models. Therefore, in Figure 2 we only present the rates for the other methods. First, we compare the proposed approach with sparse LDA. While in models 1, 2 and 8 they perform similarly, accounting for unequal covariance matrices results in drastic improvements on models 4–7. When comparing our approach to sparse QDA methods, the relative ranking often depends on $p$. For example, when $p = 100$, ridge fusion of Price et al. (2014) is better than our proposal on models 2 and 8, but is significantly worse on the same models when $p = 500$. Similarly, sparse QDA of Le and Hastie (2014) is significantly better than our proposal on models 6 and 8 when $p = 100$, but significantly worse on the same models when $p = 500$. This confirms that the proposed rule is well-suited to high-dimensional settings. Among the sparse QDA approaches, we find that the method of Li and Shao (2015) is most consistent across dimensions. In particular, it leads to better error rates on models 4 and 5 (2% difference in median error rates). Nevertheless, it still leads to significantly worse error rates on models 1, 2, 6 and 8. Finally, the proposed approach performs better than regularized discriminant analysis in all cases but model 2, $p = 100$, and performs as well or better than the sparse logistic regression in all scenarios.

Table 1: List of considered models for $\Sigma_1$ and $\Sigma_2$

| Model | $\Sigma_1$ | $\Sigma_2$ |
|---|---|---|
| 1 | equicorrelation, $b = 100$, $\rho = 0.5$ | equicorrelation, $b = 100$ $\rho = 0.5$ |
| 2 | autocorrelation, $b = 100$ $\rho = 0.8$ | equicorrelation, $b = 100$, $\rho = 0.5$ |
| 3 | autocorrelation, $b = 10$, $\rho = 0.5$ | equicorrelation, $b = 10$, $\rho = 0.8$ |
| 4 | spiked, $b = 10$ | spiked, $b = 10$ ($q_1$ and $q_2$ reversed) |
| 5 | spiked, $b = 100$ | spiked, $b = 10$ ($q_1$ and $q_2$ reversed) |
| 6 | spiked, $b = 10$ | equicorrelation, $b = 10$, $\rho = 0.8$ |
| 7 | spiked, $b = 10$ | equicorrelation, $b = 100$, $\rho = 0.3$ |
| 8 | spiked, $b = 100$ | equicorrelation, $b = 100$, $\rho = 0.3$ |

Overall, we found that no method is universally the best in terms of error rates since the relative ranking depends on the particular model and the underlying dimension. This is consistent with previous research. In the words of Wu et al. (2018), "it is difficult to imagine that there could be a universally optimal discriminant analysis method for high-dimensional data. Almost every method can enjoy some advantages under certain circumstances." Nevertheless, three methods stand out as the best across all models and dimensions: our proposal and sparse QDA methods of Le and Hastie (2014) and Li and Shao (2015). Moreover, our proposal achieves comparable, and in certain scenarios significantly better, error rates than the best other methods in all the cases with $p = 500$ except model 2.

In summary, Figure 3 shows that the proposed discriminant analysis via projections significantly improved over sparse LDA method, and results in competitive, and often better, misclassification error rates than existing QDA proposals. The real advantages of our approach, however, become certain when comparing variable selection performance and computational speed. Figure 3 reveals that the proposed method consistently uses the sparsest model (less than 50 variables for most scenarios). In comparison, the methods of Le and Hastie (2014) and Price et al. (2014) always use all $p$ variables, and are such much less interpretable.

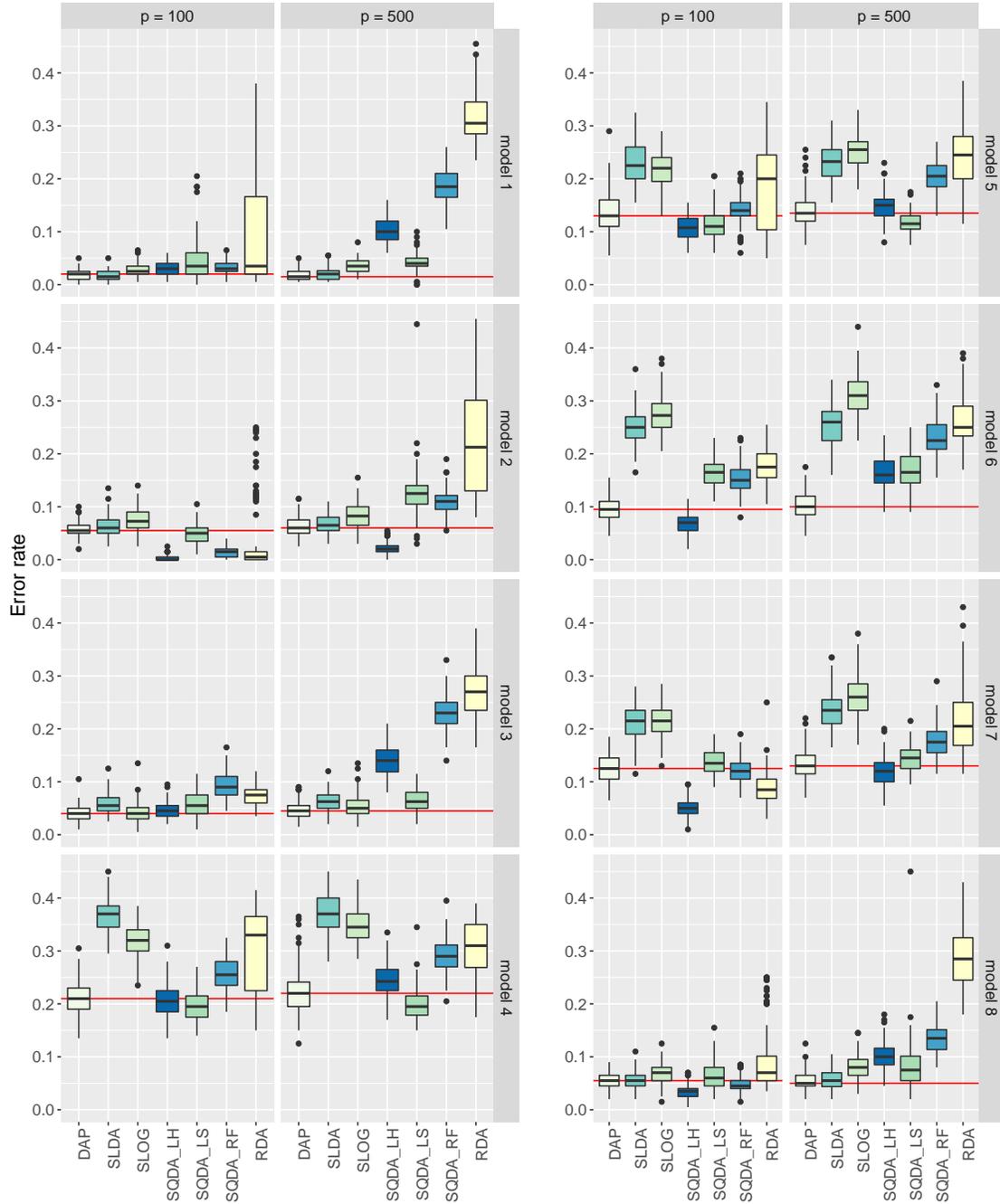

Figure 2: Misclassification error rates over 100 replications, the horizontal lines show the median errors of the proposed DAP, discriminant analysis via projections. SLDA: Sparse linear discriminant analysis; SLOG: Sparse logistic regression with interactions; SQDA_LH: Sparse QDA of Le and Hastie (2014); SQDA_LS: Sparse QDA of Li and Shao (2015); SQDA_RF: Sparse QDA via ridge fusion; RDA: Regularized discriminant analysis.

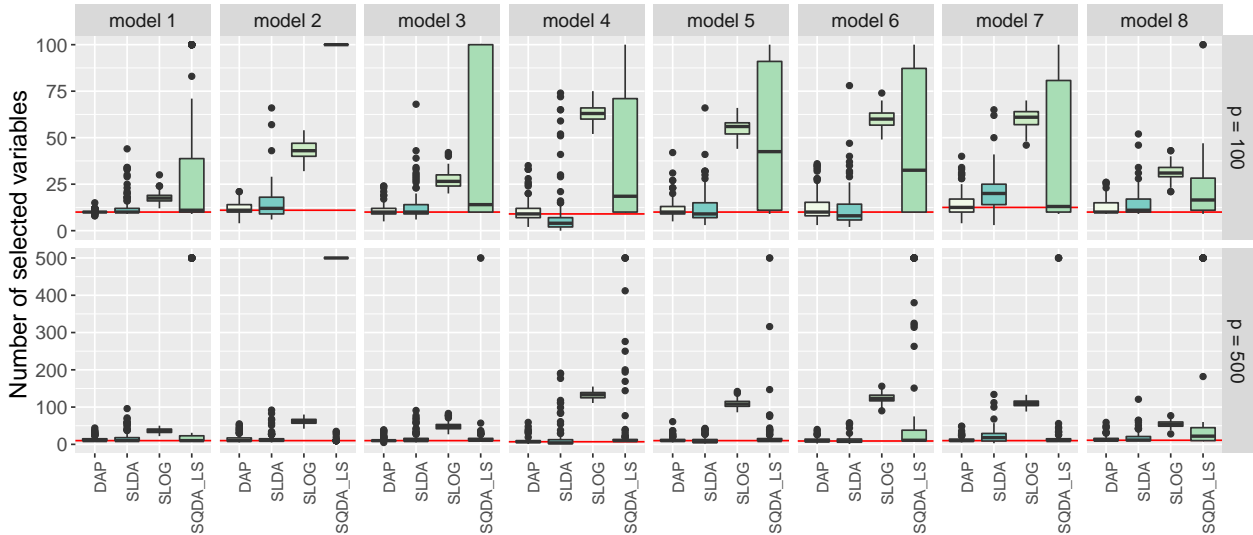

Figure 3: Number of selected variables over 100 replications, the horizontal lines indicate the median model sizes of proposed DAP, discriminant analysis via projections. RDA, SQDA_RF and SQDA_LH use all $p$ variables, not shown. SLDA: Sparse linear discriminant analysis; SLOG: Sparse logistic regression with interactions; SQDA_LH: Sparse QDA of Le and Hastie (2014); SQDA_LS: Sparse QDA of Li and Shao (2015); SQDA_RF: Sparse QDA via ridge fusion; RDA: Regularized discriminant analysis.

We further compare the execution time of each method on a Linux machine with Intel Xeon X5560 @2.80 GHz. We define execution time as the full time for method's implementation: tuning parameter selection plus model fitting plus classification. We use one instance of model 8 with $p \in \{100, 300, 500\}$, and R package `microbenchmark` (Mersmann, 2015) with 10 evaluations of each expression. Table 4.1 shows that the execution times increase dramatically with $p$ for logistic regression with interactions and sparse QDA methods, whereas the times are quite consistent across dimensions for sparse LDA, RDA and our approach. Logistic regression is noticeably faster than sparse QDA methods mainly due to the difference in tuning parameter selection criterion: it uses BIC instead of cross-validation. Using cross-validation for logistic regression makes it too computationally demanding for the range of $p$ we considered. Sparse LDA and the proposed method are the fastest, confirming that they are well-suited for the use on high-dimensional datasets in practice.

Table 2: Median time (seconds) over 10 replications to fully implement each classification method for one instance of model 8. DAP: Discriminant analysis via projections, proposed; SLDA: Sparse linear discriminant analysis; RDA: Regularized discriminant analysis; SLOG: Sparse logistic regression with interactions; SQDA_LH: Sparse QDA of Le and Hastie (2014); SQDA_RF: Sparse QDA via ridge fusion; SQDA_LS: Sparse QDA of Li and Shao (2015).

| p   | DAP | SLDA | RDA | SLOG  | SQDA_LH | SQDA_RF | SQDA_LS |
|-----|-----|------|-----|-------|---------|---------|---------|
| 100 | 0.6 | 0.4  | 3.1 | 2.7   | 139.5   | 868.5   | 52.6    |
| 300 | 1.0 | 1.4  | 5.0 | 28.8  | 2071.9  | 11681.4 | 481.5   |
| 500 | 1.4 | 1.7  | 5.0 | 117.1 | 7282.2  | 45161.7 | 1791.4  |

## 4.2 Benchmark datasets

We compare the proposed discriminant analysis via projections with competitors on three benchmark datasets: *chin* (Chin et al., 2006), *chowdary* (Chowdary et al., 2006), and *gravier* (Gravier et al., 2010). These datasets are commonly used to assess classification performance (Li and Ngom, 2013; Niu et al., 2015; Ramey et al., 2016), and are publicly available from the R package `datamicroarray` (Ramey, 2016). Below is the short description of each dataset.

*chin*: $p = 22,215$ gene expression profiles for $n = 118$ breast cancer samples with $n_1 = 75$ being ER-positive, and $n_2 = 43$ being ER-negative.

*gravier*: $p = 2,905$ gene expression profiles for $n = 168$ patients with small invasive ductal carcinomas without axillary lymph node involvement. The $n_1 = 111$ patients have no event after a 5-year diagnosis (labelled good), and $n_2 = 57$ patients have early metastasis (labelled poor).

*chowdary*: $p = 22,283$ gene expression profiles from 32 matched breast tumour tissue pairs and 20 matched colon tissue pairs leading to $n = 104$ samples with $n_1 = 64$ and $n_2 = 40$.

We randomly split each dataset 100 times preserving the class proportions, and use 80% for training and 20% for testing. To reduce the computational cost associated with sparse quadratic discriminant analysis, we reduce the number of variables at each split by selecting the top $p = 1000$ variables with largest absolute value of the two-sample t-statistic on the training data, similar approach has been taken in Cai and Liu (2011). For fair comparison, we use the same set of 1000 variables for each of the methods. We do not consider sample quadratic discriminant analysis given its uniformly poor performance in Section 4.1. We also do not consider sparse logistic regression with interactions or ridge fusion due to computational issues when $p = 1000$ and their inferiority to other approaches in Section 4.1.

The results are shown in Figure 4. For *chin* dataset, the error rates are the worst for linear discriminant analysis confirming the importance of taking into account unequal covariance matrices, and are the same for other methods. At the same time, the proposed DAP rule selects significantly smaller model than the competitors (median model size is one). For *chowdary* dataset, the best performing method is RDA (Friedman, 1989), however the relative difference is only 1 misclassification on the test data. The smallest model again corresponds to proposed DAP. For *gravier* dataset, the best performing methods are ours and sparse QDA of Le and Hastie (2014). Surprisingly, however, the method of Le and Hastie (2014) results in no variable selection on these datasets, the model size is 1000 over almost all replications (not shown). We suspect that the poor variable selection performance may be due to the crudeness of bisection procedure for selecting the tuning parameters. In summary, the proposed approach, discriminant analysis via projections, consistently selects the smallest model, often using less than 20 variables to achieve the same or better error rates than alternative methods. We conclude that it exhibits excellent prediction accuracy with the smallest model complexity.

We further analyze the *chin* dataset using variable selection results of our approach. Figure 4 reveals that the median model size is one. This means that in most of the replications it is sufficient to look at the expression level of only one gene to achieve the same misclassification error rate as the other methods. We investigate whether the same gene is selected at each replication, and find that estrogen receptor 1 gene ESR1 is selected in 97 out of 100 cases. Our finding confirms previous studies on a strong link between ESR1 gene and estrogen receptor protein expression in breast cancer patients (Holst et al., 2007; Laenkholm et al., 2012; Iwamoto et al., 2012). We refer the reader to Holst (2016) for the review on the importance of ESR1 gene amplification in breast cancer. The gene with the second highest frequency of selection, 26 out of 100 cases, is LPIN1, which is also found to be differentially expressed in ER positive and negative patients in previous studies (Chen et al., 2008). The relatively low selection frequency of LPIN1 is due to the median model size one, which leads to only ESR1 being selected and no other gene. While the strong link between ER protein expression status and ESR1 gene is not surprising, unlike the previous studies we did not focus on the ESR1 gene in advance. We consider all 22 thousand genes, and let our method determine that ESR1 is crucial for ER status of breast cancer. We want to emphasize that this insight is not possible with other approaches we tried. Regularized discriminant analysis of Friedman (1989) and sparse QDA by Le and Hastie (2014) use all 1000 variables, hence can not be directly used for identifying important genes. Sparse LDA selects a smaller number of genes, but it has worse misclassification error rate and the median model size is still 45 variables, significantly larger than the number of variables used by our approach.

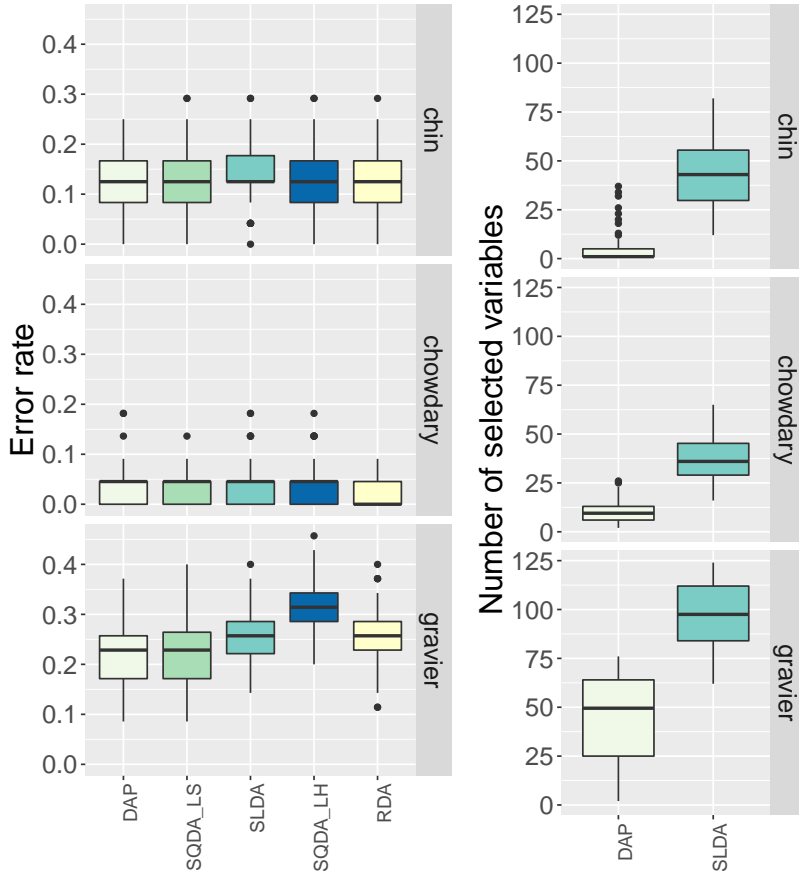

Figure 4: **Left:** Misclassification error rates over 100 splits. **Right:** Number of variables used in corresponding classification rules. DAP consistently selects the smallest model. SQDA_LS, SQDA_LH and RDA always use all $p = 1000$ variables, not shown. DAP: Discriminant analysis via projections, proposed method; SQDA_LS: Sparse QDA of Li and Shao (2015); SQDA_LH: Sparse QDA of Le and Hastie (2014); SLDA: Sparse linear discriminant analysis; RDA: Regularized discriminant analysis.

## 5 Discussion

In this work we propose a new rule for high-dimensional classification in the case of unequal covariance matrices. While the proposed approach in general differs from the Bayes rule on the population level, we show that the nonzero variables in our rule correspond to nonzero variables in the linear part of the Bayes quadratic rule. This connection combined with computational efficiency of our approach suggests that one can potentially use our method as a variable screening tool. Indeed, the empirical studies in Section 4.1 indicate that the performance of full quadratic methods deteriorates significantly with increase in $p$, however for small $p$ they are computationally feasible and may lead to better error rates. We have not explored the screening properties of our approach in this work, but leave it for future investigation.

We focus on the two-group classification setting, however extending the methodology to the multi-group setting will likely lead to even further computational gains. One of the main challenges in the multi-group case is the likely rank degeneracy of the matrix of discriminant vectors when the number of groups is large. Performing simultaneous low-rank and sparse estimation of the matrix of discriminant vectors in the multi-group case is an interesting direction for future research.

# Acknowledgement

IG was supported by NSF grant DMS-1712943. All the plots are generated using ggplot2 (Wickham, 2016).

# Appendix

## Appendix A    Implementation details

In this section we describe implementation details for the methods considered in Section 4.1. We use R package JGL (Danaher, 2013) to implement sparse QDA of Le and Hastie (2014); R package MGSDA (Gaynanova, 2016) to implement sparse LDA (Mai et al., 2012; Gaynanova et al., 2016); R package grpreg (Breheny and Huang, 2015) to implement logistic regression with pairwise interactions and lasso penalty on the vector of coefficients; R package RidgeFusion (Price, 2014) to implement ridge fusion for joint estimation of precision matrices (Price et al., 2014); R package sparsediscrim to implement regularized discriminant analysis (Friedman, 1989). We found no available R code for sparse QDA of Li and Shao (2015), and implemented the method ourselves. We use R package DAP to implement the proposed discriminant analysis via projections, which is available from the authors github page https://github.com/irinagain/DAP.

For logistic regression, we use BIC option in the grpreg to select the tuning parameter. For ridge fusion, we use the automatic selection in RidgeFusion with 5 folds. For Li and Shao (2015), we use the bisection procedure proposed in their paper with the maximal interval length set to 0.05. For all other methods, we use 5-fold cross-validation to minimize misclassification error rate.

## Appendix B    Proofs of propositions

*Proof of Proposition 2.* From Gaynanova et al. (2016), $\widetilde{v}(\lambda) = \operatorname{argmin}_v L_1(v, \lambda)$, where

$$L_1(v, \lambda) = v^{\mathrm{T}} \left(n_1 S_1 + n_2 S_2\right) v/(2n) + n_1 n_2 d^{\mathrm{T}} v v^{\mathrm{T}} d/(2n^2) - n_1^{1/2} n_2^{1/2} d^{\mathrm{T}} v/n + \lambda \|v\|_1.$$

From (6), $\{\widehat{v}_1(\lambda), \widehat{v}_2(\lambda)\} = \operatorname{argmin}_{v_1, v_2} L_2(v_1, v_2, \lambda)$, where

$$L_2(v_1, v_2, \lambda) = (v_1^{\mathrm{T}} S_1 v_1 + v_2^{\mathrm{T}} S_2 v_2)/2$$
$$+ \left(n_2 n^{-1} d^{\mathrm{T}} v_1 - 1\right)^2 / 2 + \left(n_1 n^{-1} d^{\mathrm{T}} v_2 - 1\right)^2 / 2 + \lambda \sum_{j=1}^{p} (v_{1j}^2 + v_{2j}^2)^{1/2}.$$

Under the constraint $(n/n_1)^{1/2} v_1 = (n/n_2)^{1/2} v_2 = v$, this leads to $\widehat{v}(\lambda) = \operatorname{argmin}_v L_2(v, \lambda)$, where using $c = (n_1/n)^{1/2} + (n_2/n)^{1/2}$,

$$L_2(v, \lambda) = v^{\mathrm{T}} (n_1 S_1 + n_2 S_2) v/(2n) + n_1 n_2 d^{\mathrm{T}} v v^{\mathrm{T}} d/(2n^2) - n_1^{1/2} n_2^{1/2} c d^{\mathrm{T}} v/n + \lambda \|v\|_1.$$

Furthermore,

$$L_1(v/c, \lambda/c)$$
$$= c^{-2} \left\{ v^{\mathrm{T}} (n_1 S_1 + n_2 S_2) v/(2n) + n_1 n_2 d^{\mathrm{T}} v v^{\mathrm{T}} d/(2n^2) - n_1^{1/2} n_2^{1/2} c d^{\mathrm{T}} v/n + \lambda \|v\|_1 \right\}$$
$$= c^{-2} L_2(v, \lambda).$$

Since for any $c > 0$, $\operatorname{argmin}_x f(x/c) = c\{\operatorname{argmin}_x f(x)\}$, it follows that $c \widetilde{v}(\lambda/c) = \widehat{v}(\lambda)$. □

*Proof of Proposition 3.* Since $\log(|\Sigma_2|/|\Sigma_1|)$ and $2\log(\pi_1/\pi_2)$ are present in both rules, it remains to show the equivalence of the quadratic term, the linear term and the remaining constants. Substituting $x = x - \mu + \mu$ in the Bayes rule (8) leads to

$$x^{\mathrm{T}}(\Sigma_2^{-1} - \Sigma_1^{-1}) x = (x - \mu)^{\mathrm{T}}(\Sigma_2^{-1} - \Sigma_1^{-1})(x - \mu) + 2(x - \mu)^{\mathrm{T}}(\Sigma_2^{-1} - \Sigma_1^{-1}) \mu$$
$$+ \mu^{\mathrm{T}}(\Sigma_2^{-1} - \Sigma_1^{-1}) \mu,$$
$$-2x^{\mathrm{T}}(\Sigma_2^{-1} \mu_2 - \Sigma_1^{-1} \mu_1) = -2(x - \mu)^{\mathrm{T}}(\Sigma_2^{-1} \mu_2 - \Sigma_1^{-1} \mu_1) - 2\mu^{\mathrm{T}}(\Sigma_2^{-1} \mu_2 - \Sigma_1^{-1} \mu_1).$$

From the above, the quadratic term in $(x - \mu)$ is the same as stated in the Proposition, hence it remains to consider the linear terms and the constants.

Consider the linear terms in $(x - \mu)$ from the above. Recall that $\delta = \mu_1 - \mu_2$, therefore

$$2(x - \mu)^{\mathrm{T}}(\Sigma_2^{-1} - \Sigma_1^{-1}) \mu - 2(x - \mu)^{\mathrm{T}}(\Sigma_2^{-1} \mu_2 - \Sigma_1^{-1} \mu_1)$$
$$= 2(x - \mu)^{\mathrm{T}}\{\Sigma_2^{-1}(\mu - \mu_2) - \Sigma_1^{-1}(\mu - \mu_1)\}$$
$$= 2(x - \mu)^{\mathrm{T}}(\pi_1 \Sigma_2^{-1} \delta + \pi_2 \Sigma_1^{-1} \delta),$$

which is the same as the linear term in the statement of the proposition.

Finally, we complete the proof by showing the equivalence of remaining constants.

$$\mu^{\mathrm{T}}(\Sigma_2^{-1} - \Sigma_1^{-1}) \mu - 2\mu^{\mathrm{T}}(\Sigma_2^{-1} \mu_2 - \Sigma_1^{-1} \mu_1) - \mu_1^{\mathrm{T}} \Sigma_1^{-1} \mu_1 + \mu_2^{\mathrm{T}} \Sigma_2^{-1} \mu_2$$
$$= (\mu^{\mathrm{T}} \Sigma_2^{-1} \mu - 2\mu^{\mathrm{T}} \Sigma_2^{-1} \mu_2 + \mu_2^{\mathrm{T}} \Sigma_2^{-1} \mu_2) - (\mu^{\mathrm{T}} \Sigma_1^{-1} \mu - 2\mu^{\mathrm{T}} \Sigma_1^{-1} \mu_1 + \mu_1^{\mathrm{T}} \Sigma_1^{-1} \mu_1)$$
$$= \pi_1^2 \delta^{\mathrm{T}} \Sigma_2^{-1} \delta - \pi_2^2 \delta^{\mathrm{T}} \Sigma_1^{-1} \delta.$$

□

*Proof of Proposition 4.* Since $\Psi^{\mathrm{T}}X|Y = g \sim N(\Psi^{\mathrm{T}}\mu_g, \Psi^{\mathrm{T}}\Sigma_g\Psi)$, from Proposition 3 the Bayes rule applied to $\Psi^{\mathrm{T}}x$ has the form

$$(x-\mu)^{\mathrm{T}}\Psi\left\{(\Psi^{\mathrm{T}}\Sigma_2\Psi)^{-1} - (\Psi^{\mathrm{T}}\Sigma_1\Psi)^{-1}\right\}\Psi^{\mathrm{T}}(x-\mu) + \log\left(|\Psi^{\mathrm{T}}\Sigma_2\Psi|/|\Psi^{\mathrm{T}}\Sigma_1\Psi|\right)$$
$$+ 2(x-\mu)^{\mathrm{T}}\left\{\pi_1\Psi(\Psi^{\mathrm{T}}\Sigma_2\Psi)^{-1}\Psi^{\mathrm{T}}\delta + \pi_2\Psi(\Psi^{\mathrm{T}}\Sigma_1\Psi)^{-1}\Psi^{\mathrm{T}}\delta\right\} \quad \text{(A.1)}$$
$$+ \pi_1^2\delta^{\mathrm{T}}\Psi(\Psi^{\mathrm{T}}\Sigma_2\Psi)^{-1}\Psi^{\mathrm{T}}\delta - \pi_2^2\delta^{\mathrm{T}}\Psi(\Psi^{\mathrm{T}}\Sigma_1\Psi)^{-1}\Psi^{\mathrm{T}}\delta + 2\log(\pi_1/\pi_2) > 0.$$

Since

$$(\Psi^{\mathrm{T}}\Sigma_1\Psi)^{-1} = \frac{1}{\psi_1^{\mathrm{T}}\Sigma_1\psi_1\psi_2^{\mathrm{T}}\Sigma_1\psi_2 - (\psi_1^{\mathrm{T}}\Sigma_1\psi_2)^2}\begin{pmatrix} \psi_2^{\mathrm{T}}\Sigma_1\psi_2 & -\psi_1^{\mathrm{T}}\Sigma_1\psi_2 \\ -\psi_2^{\mathrm{T}}\Sigma_1\psi_1 & \psi_1^{\mathrm{T}}\Sigma_1\psi_1 \end{pmatrix}.$$

it follows that

$$\Psi(\Psi^{\mathrm{T}}\Sigma_1\Psi)^{-1}\Psi^{\mathrm{T}} = \frac{\psi_1\psi_2^{\mathrm{T}}\Sigma_1\psi_2\psi_1^{\mathrm{T}} - \psi_2\psi_2^{\mathrm{T}}\Sigma_1\psi_1\psi_1^{\mathrm{T}} + \psi_2\psi_1^{\mathrm{T}}\Sigma_1\psi_1\psi_2^{\mathrm{T}} - \psi_1\psi_1^{\mathrm{T}}\Sigma_1\psi_2\psi_2^{\mathrm{T}}}{\psi_1^{\mathrm{T}}\Sigma_1\psi_1\psi_2^{\mathrm{T}}\Sigma_1\psi_2 - (\psi_1^{\mathrm{T}}\Sigma_1\psi_2)^2}.$$

Recall that $\psi_1 = c_1\Sigma_1^{-1}\delta$, and substituting $\delta = c_1^{-1}\Sigma_1\psi_1$ into the above equation leads to

$$\Psi(\Psi^{\mathrm{T}}\Sigma_1\Psi)^{-1}\Psi\delta = \frac{c_1^{-1}\psi_1\left\{\psi_2^{\mathrm{T}}\Sigma_1\psi_2\psi_1^{\mathrm{T}}\Sigma_1\psi_1 - (\psi_1^{\mathrm{T}}\Sigma_1\psi_2)^2\right\}}{\psi_1^{\mathrm{T}}\Sigma_1\psi_1\psi_2^{\mathrm{T}}\Sigma_1\psi_2 - (\psi_1^{\mathrm{T}}\Sigma_1\psi_2)^2} = c_1^{-1}\psi_1 = \Sigma_1^{-1}\delta.$$

Similarly, $\Psi(\Psi^{\mathrm{T}}\Sigma_2\Psi)^{-1}\Psi^{\mathrm{T}}\delta = \Sigma_2^{-1}\delta$. Substituting these into (A.1) completes the proof. □

## Appendix C  Proofs of main theorems

We will use the following quantities throughout the proofs:

$$\gamma = 1 + \max\left(\pi_1\pi_2^{-1}\|\Sigma_{1AA}^{-1/2}\Sigma_{2AA}\Sigma_{1AA}^{-1/2}\|_2, \pi_2\pi_1^{-1}\|\Sigma_{2AA}^{-1/2}\Sigma_{1AA}\Sigma_{2AA}^{-1/2}\|_2\right), \quad \text{(B.1)}$$

$$\Sigma_{gA^cA^c:A} = \Sigma_{gA^cA^c} - \Sigma_{gA^cA}\Sigma_{gAA}^{-1}\Sigma_{gA^cA} \quad (g = 1,2),$$

$$\Sigma_{d_1} = \Sigma_{1A^cA^c:A} + \pi_1\pi_2^{-1}\Big(\Sigma_{2A^cA^c} + \Sigma_{1A^cA}\Sigma_{1AA}^{-1}\Sigma_{2AA}\Sigma_{1AA}^{-1}\Sigma_{1AA^c}$$
$$- \Sigma_{1A^cA}\Sigma_{1AA}^{-1}\Sigma_{2AA^c} - \Sigma_{2A^cA}\Sigma_{1AA}^{-1}\Sigma_{1AA^c}\Big), \quad \text{(B.2)}$$

$$\Sigma_{d_2} = \Sigma_{2A^cA^c:A} + \pi_2\pi_1^{-1}\Big(\Sigma_{1A^cA^c} + \Sigma_{2A^cA}\Sigma_{2AA}^{-1}\Sigma_{1AA}\Sigma_{2AA}^{-1}\Sigma_{2AA^c}$$
$$- \Sigma_{2A^cA}\Sigma_{2AA}^{-1}\Sigma_{1AA^c} - \Sigma_{1A^cA}\Sigma_{2AA}^{-1}\Sigma_{2AA^c}\Big).$$

The quantities in (B.2) can be viewed as conditional variance terms, their origin is made precise in Lemma 2. Let $\sigma^2_{gjj:A} = e_j^{\mathrm{T}}\Sigma_{gA^cA^c:A}e_j$ and $\sigma^2_{jdg} = e_j^{\mathrm{T}}\Sigma_{d_g}e_j$ be the diagonal elements of corresponding matrices. Under Assumption 4, $\sigma_{gjj:A}$, $\sigma_{jdg}$ and $\gamma$ can be treated as constants.

We define the oracle $(\tilde{v}_{1A}, \tilde{v}_{2A})$ as the solution to

$$\underset{v_1, v_2 \in \mathbb{R}^s}{\text{minimize}}\left\{n_1^{-1}\|X_{1A}v_1 - 1_{n_1}\|_2^2/2 + n_2^{-1}\|X_{2A}v_2 + 1_{n_2}\|_2^2/2 + \lambda\sum_{j=1}^{s}(v_{1j}^2 + v_{2j}^2)^{1/2}\right\}, \quad \text{(B.3)}$$

and let $\tilde{u}_A = (\tilde{u}_{1A}, \tilde{u}_{2A})$ be the subgradient of $\sum_{j=1}^{s}(v_{1j}^2 + v_{2j}^2)^{1/2}$ evaluated at $(\tilde{v}_{1A}, \tilde{v}_{2A})$

$$\tilde{u}_{Aj} = \begin{cases} \tilde{v}_{Aj}/\|\tilde{v}_{Aj}\|_2, & \text{if } \|\tilde{v}_{Aj}\|_2 \neq 0; \\ \in \{u : \|u\|_2 \leq 1\}, & \text{if } \|\tilde{v}_{Aj}\|_2 = 0. \end{cases} \quad (B.4)$$

**Theorem 3** (Equivalent to Theorem 1). *Let Assumptions 1–3 hold. Let the sample sizes satisfy*

$$\min(n_1, n_2) \gtrsim \max_{g=1,2} \|\Sigma_{gAA}^{-1}\|_2 \max_{g=1,2;\ j \in A^c}(\sigma_{gjj:A}^2 \vee \sigma_{jdg}^2) s \log\{(p-s)\eta^{-1}\},$$

*for some $\eta \in (0,1)$, and the tuning parameter satisfy*

$$\lambda \gtrsim \max_{g=1,2;\ j \in A^c}(\sigma_{gjj:A}^2 \vee \sigma_{jdg}^2)\left[n^{-1}\log\{(p-s)\eta^{-1}\}\right]^{1/2}.$$

*Then $\mathrm{pr}(\widehat{A} \subseteq A) \geq 1 - \eta$.*

*Proof.* Using the results of Section 2.3,

$$[\widehat{v}_1\ \widehat{v}_2] = \operatorname*{argmin}_{v_1 \in \mathbb{R}^p, v_2 \in \mathbb{R}^p} \left\{\widehat{L}_{\psi_1}(v_1) + \widehat{L}_{\psi_2}(v_2) + \lambda \sum_{j=1}^{p}(v_{1j}^2 + v_{2j}^2)^{1/2}\right\},$$

$$2\{\widehat{L}_{\psi_1}(v_1) + \widehat{L}_{\psi_2}(v_2)\} = v_1^\mathrm{T} S_1 v_1 + v_2^\mathrm{T} S_2 v_2 + (n^{-1}n_2 d^\mathrm{T} v_1 - 1)^2 + (n^{-1}n_1 d^\mathrm{T} v_2 - 1)^2.$$

Let $\rho_1 = n_1/n$ and $\rho_2 = n_2/n$. The optimality conditions (Boyd and Vandenberghe, 2004, Chapter 5) lead to

$$(S_{1AA} + \rho_2^2 d_A d_A^\mathrm{T})\widehat{v}_{1A} + (S_{1AA^c} + \rho_2^2 d_A d_{A^c}^\mathrm{T})\widehat{v}_{1A^c} - \rho_2 d_A = -\lambda u_{1A},$$
$$(S_{2AA} + \rho_1^2 d_A d_A^\mathrm{T})\widehat{v}_{2A} + (S_{2AA^c} + \rho_1^2 d_A d_{A^c}^\mathrm{T})\widehat{v}_{2A^c} - \rho_1 d_A = -\lambda u_{2A},$$
$$(S_{1A^cA} + \rho_2^2 d_{A^c} d_A^\mathrm{T})\widehat{v}_{1A} + (S_{1A^cA^c} + \rho_2^2 d_{A^c} d_{A^c}^\mathrm{T})\widehat{v}_{1A^c} - \rho_2 d_{A^c} = -\lambda u_{1A^c},$$
$$(S_{2A^cA} + \rho_1^2 d_{A^c} d_A^\mathrm{T})\widehat{v}_{2A} + (S_{2A^cA^c} + \rho_1^2 d_{A^c} d_{A^c}^\mathrm{T})\widehat{v}_{2A^c} - \rho_2 d_{A^c} = -\lambda u_{2A^c},$$

where $u$ is defined in (7). Consider $\widehat{v}_1 = (\tilde{v}_{1A}, 0_{p-s})$, $\widehat{v}_2 = (\tilde{v}_{2A}, 0_{p-s})$, where $\tilde{v}_{1A}$, $\tilde{v}_{2A}$ are the solutions to the oracle problem (B.3). From the above optimality conditions, it is sufficient to have

$$\left\|(S_{1A^cA} + \rho_2^2 d_{A^c} d_A^\mathrm{T})\tilde{v}_{1A} - \rho_2 d_{A^c},\ (S_{2A^cA} + \rho_1^2 d_{A^c} d_A^\mathrm{T})\tilde{v}_{2A} - \rho_1 d_{A^c}\right\|_{\infty,2} < \lambda$$

for $\widehat{V} = [\widehat{v}_1\ \widehat{v}_2]$ to be the solution to (6), which leads to $\widehat{A} \subseteq A$. We next show that the above inequality holds with high probability under the stated conditions.

Using the form of $\tilde{v}_{1A}$ (Theorem 5) and Sherman–Morrison identity,

$$(S_{1A^cA} + \rho_2^2 d_{A^c} d_A^\mathrm{T})\tilde{v}_{1A} - \rho_2 d_{A^c}$$
$$= S_{1A^cA}\rho_2 S_{1AA}^{-1} d_A(1 + \rho_2^2 d_A^\mathrm{T} S_{1AA}^{-1} d_A)^{-1} + \rho_2^2 d_{A^c} d_A \rho_2^\mathrm{T} S_{1AA}^{-1} d_A(1 + \rho_2^2 d_A^\mathrm{T} S_{1AA}^{-1} d_A)^{-1}$$
$$\quad - \lambda S_{1A^cA}\left(S_{1AA} + \rho_2^2 d_A d_A^\mathrm{T}\right)^{-1}\tilde{u}_{1A} - \lambda \rho_2^2 d_{A^c} d_A^\mathrm{T}\left(S_{1AA} + \rho_2^2 d_A d_A^\mathrm{T}\right)^{-1}\tilde{u}_{1A} - \rho_2 d_{A^c}$$
$$= \rho_2\left(S_{1A^cA} S_{1AA}^{-1} d_A - d_{A^c}\right)(1 + \rho_2^2 d_A^\mathrm{T} S_{1AA}^{-1} d_A)^{-1} - \lambda S_{1A^cA} S_{1AA}^{-1}\tilde{u}_{1A}$$
$$\quad + \lambda\rho_2^2 S_{1A^cA} S_{1AA}^{-1} d_A d_A^\mathrm{T} S_{1AA}^{-1}\tilde{u}_{1A}(1 + \rho_2^2 d_A^\mathrm{T} S_{1AA}^{-1} d_A)^{-1}$$
$$\quad - \lambda \rho_2^2 d_{A^c} d_A^\mathrm{T} S_{1AA}^{-1}\tilde{u}_{1A}(1 + \rho_2^2 d_A^\mathrm{T} S_{1AA}^{-1} d_A)^{-1}$$
$$= \rho_2\left(S_{1A^cA} S_{1AA}^{-1} d_A - d_{A^c}\right)(1 + \rho_2^2 d_A^\mathrm{T} S_{1AA}^{-1} d_A)^{-1} - \lambda S_{1A^cA} S_{1AA}^{-1}\tilde{u}_{1A}$$
$$\quad + \rho_2^2 \lambda(S_{1A^CA} S_{1AA}^{-1} d_A - d_{A^c}) d_A^\mathrm{T} S_{1AA}^{-1}\tilde{u}_{1A}(1 + \rho_2^2 d_A^\mathrm{T} S_{1AA}^{-1} d_A)^{-1}.$$

Using normality, there exist $U_1 \in \mathbb{R}^{p \times (n_1-1)}$ with columns $u_{1,i} \sim \mathcal{N}(0, \Sigma_1)$ such that $(n_1 - 1)S_1 = U_1 U_1^T$. Let $E_{d1} = d_{A^c} - \Sigma_{1A^cA}\Sigma_{1AA}^{-1}d_A$, $E_{U1} = U_{1A^c} - \Sigma_{1A^cA}\Sigma_{1AA}^{-1}U_{1A}$. Then

$$
\begin{aligned}
S_{1A^cA}S_{1AA}^{-1} &= (n_1 - 1)^{-1}U_{1A^c}U_{1A}^T S_{1AA}^{-1} \\
&= (n_1 - 1)^{-1}E_{U1}U_{1A}^T S_{1AA}^{-1} + (n_1 - 1)^{-1}\Sigma_{1A^cA}\Sigma_{1AA}^{-1}U_{1A}U_{1A}^T S_{1,AA}^{-1} \\
&= \Sigma_{1A^cA}\Sigma_{1AA}^{-1} + (n_1 - 1)^{-1}E_{U1}U_{1A}^T S_{1AA}^{-1},
\end{aligned}
$$

and $S_{1A^cA}S_{1AA}^{-1}d_A - d_{A^c} = (n_1 - 1)^{-1}E_{U1}U_{1A}^T S_{1AA}^{-1}d_A - E_{d1}$. Combining the above two displays gives

$$
\begin{aligned}
&(S_{1A^cA} + \rho_2^2 d_{A^c}d_A^T)\tilde{v}_{1A} - \rho_2 d_{A^c} \\
&= -\lambda\Sigma_{1A^cA}\Sigma_{1AA}^{-1}\tilde{u}_{1A} - \lambda(n_1 - 1)^{-1}E_{U1}U_{1A}^T S_{1AA}^{-1}\tilde{u}_{1A} \\
&\quad + (n_1 - 1)^{-1}E_{U1}U_{1A}^T S_{1AA}^{-1}d_A\rho_2(1 + \rho_2^2 d_A^T S_{1AA}^{-1}d_A)^{-1} - E_{d1}\rho_2(1 + \rho_2^2 d_A^T S_{1AA}^{-1}d_A)^{-1} \\
&\quad + \lambda(n_1 - 1)^{-1}E_{U1}U_{1A}^T S_{1AA}^{-1}d_A\rho_2^2 d_A^T S_{1AA}^{-1}\tilde{u}_{1A}(1 + \rho_2^2 d_A^T S_{1AA}^{-1}d_A)^{-1} \\
&\quad - \lambda E_{d1}\rho_2^2 d_A^T S_{1AA}^{-1}\tilde{u}_{1A}(1 + \rho_2^2 d_A^T S_{1AA}^{-1}d_A)^{-1} \\
&= -\lambda\Sigma_{1A^cA}\Sigma_{1AA}^{-1}\tilde{u}_{1A} + (n_1 - 1)^{-1}E_{U1}U_{1A}^T S_{1AA}^{-1}d_A\rho_2(1 + \rho_2^2 d_A^T S_{1AA}^{-1}d_A)^{-1} \\
&\quad - E_{d1}\rho_2(1 + \rho_2^2 d_A^T S_{1AA}^{-1}d_A)^{-1} - \lambda E_{d1}\rho_2^2 d_A^T S_{1AA}^{-1}\tilde{u}_{1A}(1 + \rho_2^2 d_A^T S_{1AA}^{-1}d_A)^{-1} \\
&\quad - \lambda(n_1 - 1)^{-1}E_{U1}U_{1A}^T S_{1AA}^{-1}(I + \rho_2^2 d_A d_A^T S_{1AA}^{-1})^{-1}\tilde{u}_{1A}.
\end{aligned}
$$

Similarly,

$$
\begin{aligned}
&(S_{2A^cA} + \rho_1^2 d_{A^c}d_A^T)\tilde{v}_{2A} - \rho_1 d_{A^c} \\
&= -\lambda\Sigma_{2A^cA}\Sigma_{2AA}^{-1}\tilde{u}_{2A} + (n_2 - 1)^{-1}E_{U2}U_{2A}^T S_{2AA}^{-1}d_A\rho_1(1 + \rho_1^2 d_A^T S_{2AA}^{-1}d_A)^{-1} \\
&\quad - E_{d2}\rho_1(1 + \rho_1^2 d_A^T S_{2AA}^{-1}d_A)^{-1} - \lambda E_{d2}\rho_1^2 d_A^T S_{2AA}^{-1}\tilde{u}_{2A}(1 + \rho_1^2 d_A^T S_{2AA}^{-1}d_A)^{-1} \\
&\quad - \lambda(n_2 - 1)^{-1}E_{U2}U_{2A}^T S_{2AA}^{-1}(I + \rho_1^2 d_A d_A^T S_{2AA}^{-1})^{-1}\tilde{u}_{2A}.
\end{aligned}
$$

Therefore, using triangle inequality,

$$
\begin{aligned}
&\left\|(S_{1A^cA} + \rho_2^2 d_{A^c}d_A^T)\tilde{v}_{1A} - \rho_2 d_{A^c} \,,\, (S_{2A^cA} + \rho_1^2 d_{A^c}d_A^T)\tilde{v}_{2A} - \rho_1 d_{A^c}\right\|_{\infty,2} \\
&\leq \lambda\|\Sigma_{1A^cA}\Sigma_{1AA}^{-1}\tilde{u}_{1A}, \Sigma_{2A^cA}\Sigma_{2AA}^{-1}\tilde{u}_{2A}\|_{\infty,2} + I_1 + I_2 + I_3 + I_4,
\end{aligned}
$$

where

$$
\begin{aligned}
I_1 &= \|\rho_2(1 + \rho_2^2 d_A^T S_{1AA}^{-1}d_A)^{-1}E_{d1}, \rho_1(1 + \rho_1^2 d_A^T S_{2AA}^{-1}d_A)^{-1}E_{d2}\|_{\infty,2}, \\
I_2 &= \left\|(n_1 - 1)^{-1}\frac{\rho_2 E_{U1}U_{1A}^T S_{1AA}^{-1}d_A}{1 + \rho_2^2 d_A^T S_{1AA}^{-1}d_A}, (n_2 - 1)^{-1}\frac{\rho_1 E_{U2}U_{2A}^T S_{2AA}^{-1}d_A}{1 + \rho_1^2 d_A^T S_{2AA}^{-1}d_A}\right\|_{\infty,2}, \\
I_3 &= \left\|\frac{E_{U1}U_{1A}^T S_{1AA}^{-1}}{n_1 - 1}(I + \rho_2^2 d_A d_A^T S_{1AA}^{-1})^{-1}\tilde{u}_{1A}, \frac{E_{U2}U_{2A}^T S_{2AA}^{-1}}{n_2 - 1}(I + \rho_1^2 d_A d_A^T S_{2AA}^{-1})^{-1}\tilde{u}_{2A}\right\|_{\infty,2}, \\
I_4 &= \left\|\frac{\rho_2^2}{1 + \rho_2^2 d_A^T S_{1AA}^{-1}d_A}E_{d1}d_A^T S_{1AA}^{-1}\tilde{u}_{1A}, \frac{\rho_1^2}{1 + \rho_1^2 d_A^T S_{2AA}^{-1}d_A}E_{d2}d_A^T S_{2AA}^{-1}\tilde{u}_{2A}\right\|_{\infty,2}.
\end{aligned}
$$

By the irrepresentability condition (Assumption 3), there exist $\alpha \in (0, 1]$ such that

$$\|\Sigma_{1A^cA}\Sigma_{1AA}^{-1}\tilde{u}_{1A}, \Sigma_{2A^cA}\Sigma_{2AA}^{-1}\tilde{u}_{2A}\|_{\infty,2} \leq 1 - \alpha.$$

To conclude the proof, it is sufficient to show that with probability at least $1 - \eta$ each $I_k \leq \lambda\alpha/4$, $k = 1, \ldots, 4$. Next, we consider each of these four terms separately.

1. Show $I_1 \leq \lambda\alpha/4$ with probability at least $1 - \eta/4$. By Lemma 2, $e_j^\mathrm{T} E_{dg} \sim \mathcal{N}(0, \sigma_{jdg}^2/n_g)$. Applying standard normal concentration inequality, there exist constant $C > 0$ such that

$$\mathrm{pr}\Big(\bigcap_{j \in A^c} \big\{|e_j^\mathrm{T} E_{dg}| \geq C \max_{j \in A^c} \sigma_{jdg}\big[n_g^{-1} \log\{(p-s)\eta^{-1}\}\big]^{1/2}\big\}\Big) \leq \eta/4.$$

Since

$$\|\rho_2(1 + \rho_2^2 d_A^\mathrm{T} S_{1AA}^{-1} d_A)^{-1} E_{d1}, \rho_1(1 + \rho_1^2 d_A^\mathrm{T} S_{1AA}^{-1} d_A)^{-1} E_{d2}\|_{\infty,2}$$
$$\leq \sqrt{2} \max\big\{\rho_2(1 + \rho_2^2 d_A^\mathrm{T} S_{1AA}^{-1} d_A)^{-1} \|E_{d1}\|_\infty, \rho_1(1 + \rho_1^2 d_A^\mathrm{T} S_{1AA}^{-1} d_A)^{-1} \|E_{d2}\|_\infty\big\}$$
$$\leq \sqrt{2} \max(\|E_{d1}\|_\infty, \|E_{d2}\|_\infty),$$

it follows that there exist constant $C > 0$ such that

$$\mathrm{pr}\Big(I_1 \geq C \max_{g=1,2;\ j \in A^c} \sigma_{jdg}\big[\log\{(p-s)\eta^{-1}\}/\min(n_1, n_2)\big]^{1/2}\Big) \leq \eta/4.$$

Therefore, $I_1 \leq \lambda\alpha/4$ with probability at least $1 - \eta/4$ under the conditions of the theorem.

2. Show $I_2 \leq \lambda\alpha/4$ with probability at least $1 - \eta/4$. By Lemma 2, $E_{Ug} \sim \mathcal{N}(0, \Sigma_{gA^c A^c:A} \otimes I_{n_g-1})$ for $g = 1, 2$, and is independent of $U_{gA}$ and $d$. Hence,

$$\rho_2(1 + \rho_2^2 d_A^\mathrm{T} S_{1AA}^{-1} d_A)^{-1} e_j^\mathrm{T}(n_1 - 1)^{-1} E_{U1} U_{1A}^\mathrm{T} S_{1AA}^{-1} d_A | U_{1A}, d_A$$
$$\sim \mathcal{N}\big\{0, \sigma_{1jj:A}^2 (n_1 - 1)^{-1} \rho_2^2 d_A^\mathrm{T} S_{1AA}^{-1} d_A (1 + \rho_2^2 d_A^\mathrm{T} S_{1AA}^{-1} d_A)^{-2}\big\}.$$

Define $L = (1 + \rho_2^2 d_A^\mathrm{T} S_{1AA}^{-1} d_A)^{-2} \rho_2^2 d_A^\mathrm{T} S_{1AA}^{-1} d_A$. Using standard normal concentration inequality, there exist constant $C > 0$ such that conditionally on $L$, the event

$$\bigcap_{j \in A^c} \big\{\rho_2(1 + \rho_2^2 d_A^\mathrm{T} S_{1AA}^{-1} d_A)^{-1} |e_j^\mathrm{T}(n_1 - 1)^{-1} E_{U1} U_{1A}^\mathrm{T} S_{1AA}^{-1} d_A|$$
$$\geq C \max_{j \in A^c} \sigma_{1jj:A}\big[L n_1^{-1} \log\{(p-s)\eta^{-1}\}\big]^{1/2}\big\}$$

has probability at most $\eta/4$. Since $L = (1 + \rho_2^2 d_A^\mathrm{T} S_{1AA}^{-1} d_A)^{-2} \rho_2^2 d_A^\mathrm{T} S_{1AA}^{-1} d_A \leq (1 + \rho_2^2 d_A^\mathrm{T} S_{1AA}^{-1} d_A)^{-1} \leq 1$, it follows that with probability at least $1 - \eta/4$

$$\frac{\rho_2}{1 + \rho_2^2 d_A^\mathrm{T} S_{1AA}^{-1} d_A} \Big\|\frac{E_{U1} U_{1A}^\mathrm{T} S_{1AA}^{-1} d_A}{n_1 - 1}\Big\|_\infty \leq C\big[\max_{j \in A^c} \sigma_{1jj:A} n_1^{-1} \log\{(p-s)\eta^{-1}\}\big]^{1/2}.$$

The case $g = 2$ is similar, leading to the desired bound under the conditions of the theorem.

3. Show $I_3 \leq \alpha/4$ with probability at least $1 - \eta/4$. Similar to part 2,

$$e_j^\mathrm{T}(n_1 - 1)^{-1} E_{U1} U_{1A}^\mathrm{T} S_{1AA}^{-1} (I + \rho_2^2 d_A d_A^\mathrm{T} S_{1AA}^{-1})^{-1} \tilde{u}_{1A} | U_{1A}, \tilde{u}_{1A}, d_A$$
$$\sim \mathcal{N}\big(0, (n_1 - 1)^{-1} \sigma_{1jj:A}^2 \tilde{u}_{1A}^\mathrm{T}(S_{1AA} + \rho_2^2 d_A d_A^\mathrm{T})^{-1} S_{1AA}(S_{1AA} + \rho_2^2 d_A d_A^\mathrm{T})^{-1} \tilde{u}_{1A}\big).$$

Define $L = \tilde{u}_{1A}^{\mathrm{T}}(S_{1AA} + \rho_2^2 d_A d_A^{\mathrm{T}})^{-1} S_{1AA}(S_{1AA} + \rho_2^2 d_A d_A^{\mathrm{T}})^{-1} \tilde{u}_{1A}$. As in part 2, there exist constant $C > 0$ such that conditionally on $L$ the event

$$\bigcap_{j \in A^c} \left\{ |e_j^{\mathrm{T}}(n_1 - 1)^{-1} E_{U1} U_{1A}^{\mathrm{T}} S_{1AA}^{-1} (I + \rho_2^2 d_A d_A^{\mathrm{T}} S_{1AA}^{-1})^{-1} \tilde{u}_{1A}| \right.$$

$$\left. \geq C \max_{j \in A^c} \sigma_{1jj:A} \left[ L n_1^{-1} \log\{(p-s)\eta^{-1}\} \right]^{1/2} \right\}$$

has probability at most $\eta/4$. Furthermore,

$$L \leq \|\tilde{u}_{1A}\|_2^2 \|(S_{1AA} + \rho_2^2 d_A d_A^{\mathrm{T}})^{-1} S_{1AA}(S_{1AA} + \rho_2^2 d_A d_A^{\mathrm{T}})^{-1}\|_2$$
$$\leq s \|S_{1AA}^{-1/2}(I + \rho_2^2 S_{1AA}^{-1/2} d_A d_A^{\mathrm{T}} S_{1AA}^{-1/2})^{-2} S_{1AA}^{-1/2}\|_2^2$$
$$\leq s \|S_{1AA}^{-1}\|_2,$$

where in the last inequality we used $\|\tilde{u}_{1A}\|_2^2 + \|\tilde{u}_{2A}\|_2^2 \leq s$ by definition of subgradient. By Lemma 3, there exist constant $C > 0$ such that with probability at least $1 - \eta/4$

$$\|S_{1AA}^{-1}\|_2 \leq \|\Sigma_{1AA}^{-1}\|_2 \left[ 1 + C\{n_1^{-1} \log(\eta^{-1})\}^{1/2} \right].$$

Combining the above displays leads to

$$\|(n_1 - 1)^{-1} E_{U1} U_{1A}^{\mathrm{T}} S_{1AA}^{-1} (I + \rho_2^2 d_A d_A^{\mathrm{T}} S_{1AA}^{-1})^{-1} \tilde{u}_{gA}\|_\infty$$
$$\leq C \max_{j \in A^c} \sigma_{1jj:A} \left[ \|\Sigma_{1AA}^{-1}\|_2 n_1^{-1} s \log\{(p-s)\eta^{-1}\} \right]^{1/2}$$

with probability at least $1 - \eta/4$. The proof for $g = 2$ is similar leading to the desired bound.

4. Show $I_4 \leq \alpha/4$ with probability at least $1 - \eta/4$.

By Lemma 2, $e_j^{\mathrm{T}} E_{dg} \sim \mathcal{N}(0, n_g^{-1} \sigma_{jdg}^2)$, where $\sigma_{jdg}$ is from Lemma 2. Then

$$\rho_2^2 (1 + \rho_2^2 d_A^{\mathrm{T}} S_{1AA}^{-1} d_A)^{-1} e_j^{\mathrm{T}} E_{d1} d_A^{\mathrm{T}} S_{1AA}^{-1} \tilde{u}_{1A} | U_{1A}, \tilde{u}_{1A}, d_A$$
$$\sim \mathcal{N}\left(0, \frac{\sigma_{jd1}^2 \rho_2^4}{n_1(1 + \rho_2^2 d_A^{\mathrm{T}} S_{1AA}^{-1} d_A)^2} \tilde{u}_{1A}^{\mathrm{T}} S_{1AA}^{-1} d_A d_A^{\mathrm{T}} S_{1AA}^{-1} \tilde{u}_{1A}\right).$$

Define $L = (1 + \rho_2^2 d_A^{\mathrm{T}} S_{1AA}^{-1} d_A)^{-2} \rho_2^4 \tilde{u}_{1A}^{\mathrm{T}} S_{1AA}^{-1} d_A d_A^{\mathrm{T}} S_{1AA}^{-1} \tilde{u}_{1A}$. Using standard normal concentration inequality there exist constant $C > 0$ such that conditionally on $L$ the event

$$\bigcap_{j \in A^c} \left\{ \frac{\rho_2^2}{1 + \rho_2^2 d_A^{\mathrm{T}} S_{1AA}^{-1} d_A} e_j^{\mathrm{T}} E_{d1} d_A^{\mathrm{T}} S_{1AA}^{-1} \tilde{u}_{1A} \geq C \max_{j \in A^c} \sigma_{jd1} \left[ L n_1^{-1} \log\{(p-s)\eta^{-1}\} \right]^{1/2} \right\}$$

has probability at most $\eta/4$. Furthermore,

$$L = (1 + \rho_2^2 d_A^{\mathrm{T}} S_{1AA}^{-1} d_A)^{-2} \rho_2^4 (\tilde{u}_{1A}^{\mathrm{T}} S_{1AA}^{-1/2} S_{1AA}^{-1/2} d_A)^2$$
$$\leq \rho_2^2 (1 + \rho_2^2 d_A^{\mathrm{T}} S_{1AA}^{-1} d_A)^{-2} \rho_2^2 d_A^{\mathrm{T}} S_{1AA}^{-1} d_A \tilde{u}_{1A}^{\mathrm{T}} S_{1AA}^{-1} \tilde{u}_{1A}$$
$$\leq \rho_2^2 \tilde{u}_{1A}^{\mathrm{T}} S_{1AA}^{-1} \tilde{u}_{1A}$$
$$\leq s \|S_{1AA}^{-1}\|_2,$$

where in the last inequality we used $\|\tilde{u}_{1A}\|_2^2 + \|\tilde{u}_{2A}\|_2^2 \leq s$ by definition of subgradient. Similar to part 3, this means that there exists constant $C > 0$ such that

$$\left\|\frac{\rho_2^2}{1 + \rho_2^2 d_A^T S_{1AA}^{-1} d_A} E_{d1} d_A^T S_{1AA}^{-1} \tilde{u}_{1A}\right\|_\infty \geq C \max_{j \in A^c} \sigma_{jd1} \left[\|\Sigma_{1AA}^{-1}\|_2 n_1^{-1} s \log\{(p-s)\eta^{-1}\}\right]^{1/2}$$

with probability at most $\eta/4$. The proof for $g = 2$ is analogous, leading to the desired bound. $\square$

**Theorem 4** (Equivalent to Theorem 2). *Assume the conditions of Theorem 3 hold. If in addition $\psi_{\min} \gtrsim \lambda s^{1/2} \max_g \|\Sigma_{g,AA}^{-1}\|_2 (\max_g \delta_A^T \Sigma_{gAA}^{-1} \delta_A \vee \gamma)$, then $\mathrm{pr}(\widehat{A} = A) \geq 1 - \eta$.*

*Proof of Theorem 4.* Consider the oracle solution

$$\tilde{v}_{1A} = \rho_2 S_{1AA}^{-1} d_A (1 + \rho_2^2 d_A^T S_{1AA}^{-1} d_A)^{-1} - \lambda \left(S_{1AA} + \rho_2^2 d_A d_A^T\right)^{-1} \tilde{u}_{1A},$$
$$\tilde{v}_{2A} = \rho_1 S_{2AA}^{-1} d_A (1 + \rho_1^2 d_A^T S_{2AA}^{-1} d_A)^{-1} - \lambda \left(S_{2AA} + \rho_1^2 d_A d_A^T\right)^{-1} \tilde{u}_{2A};$$

where $\tilde{u}_A$ is defined in (B.4). To show $\widehat{A} = A$, it is sufficient to show

$$\min_{j \in A} \left\|\rho_2(1 + \rho_2^2 d_A^T S_{1AA}^{-1} d_A)^{-1} e_j^T S_{1AA}^{-1} d_A, \; \rho_1(1 + \rho_1^2 d_A^T S_{2AA}^{-1} d_A)^{-1} e_j^T S_{2AA}^{-1} d_A\right\|_2$$
$$\geq \lambda \max_{j \in A} \|e_j^T \left(S_{1AA} + \rho_2^2 d_A d_A^T\right)^{-1} \tilde{u}_{1A}, \; e_j^T \left(S_{2AA} + \rho_1^2 d_A d_A^T\right)^{-1} \tilde{u}_{2A}\|_2.$$
(B.5)

Consider the right-hand side in (B.5)

$$\max_{j \in A} \|e_j^T \left(S_{1AA} + \rho_2^2 d_A d_A^T\right)^{-1} \tilde{u}_{1A}, \; e_j^T \left(S_{2AA} + \rho_1^2 d_A d_A^T\right)^{-1} \tilde{u}_{2A}\|_2$$
$$= \max_{j \in A} \left[\left\{e_j^T \left(S_{1AA} + \rho_2^2 d_A d_A^T\right)^{-1} \tilde{u}_{1A}\right\}^2 + \left\{e_j^T \left(S_{2AA} + \rho_1^2 d_A d_A^T\right)^{-1} \tilde{u}_{2A}\right\}^2\right]^{1/2}$$
$$\leq \max_{j \in A} \left\{\|e_j^T \left(S_{1AA} + \rho_2^2 d_A d_A^T\right)^{-1}\|_2^2 \|\tilde{u}_{1A}\|_2^2 + \|e_j^T \left(S_{2AA} + \rho_1^2 d_A d_A^T\right)^{-1}\|_2^2 \|\tilde{u}_{2A}\|_2^2\right\}^{1/2}$$
$$\leq \max_{j \in A} \left\{\|e_j^T \left(S_{1AA} + \rho_2^2 d_A d_A^T\right)^{-1}\|_2 \vee \|e_j^T \left(S_{2AA} + \rho_1^2 d_A d_A^T\right)^{-1}\|_2\right\} \left(\|\tilde{u}_{1A}\|_2^2 + \|\tilde{u}_{2A}\|_2^2\right)^{1/2}$$
$$\leq \left\{\|(S_{1AA} + \rho_2^2 d_A d_A^T)^{-1}\|_2 \vee \|(S_{2AA} + \rho_1^2 d_A d_A^T)^{-1}\|_2\right\} s^{1/2}.$$

Furthermore,

$$\|\left(S_{1AA} + \rho_2^2 d_A d_A^T\right)^{-1}\|_2 = \|S_{1AA}^{-1/2} \left(I + \rho_2^2 S_{1AA}^{-1/2} d_A d_A^T S_{1AA}^{-1/2}\right)^{-1} S_{1AA}^{-1/2}\|_2 \leq \|S_{1AA}^{-1}\|_2,$$

and similarly $\|(S_{2AA} + \rho_1^2 d_A d_A^T)^{-1}\|_2 \leq \|S_{2AA}^{-1}\|_2$. Using Lemma 3

$$\max_{j \in A} \|e_j^T \left(S_{1AA} + \rho_2^2 d_A d_A^T\right)^{-1} \tilde{u}_{1A}, \; e_j^T \left(S_{2AA} + \rho_1^2 d_A d_A^T\right)^{-1} \tilde{u}_{2A}\|_2$$
$$\leq \max_g \|\Sigma_{gAA}^{-1}\|_2 s^{1/2} \left[1 + C\{s \log(\eta^{-1})/\min(n_1, n_2)\}^{1/2}\right]$$

with probability at least $1 - \eta$.

Consider the left-hand side in (B.5). Applying Lemma 1 and Corollary 1, there exist constants $C_1$, $C_2$ such that with probability at least $1 - \eta$

$$\min_{j \in A} \left\| \rho_2 (1 + \rho_2^2 d_A^T S_{1AA}^{-1} d_A)^{-1} e_j^T \Sigma_{1AA}^{-1} \delta_A, \; \rho_1 (1 + \rho_1^2 d_A^T S_{2AA}^{-1} d_A)^{-1} e_j^T \Sigma_{2AA}^{-1} \delta_A \right\|_2$$

$$\geq \left[ 1 + C_1 \max_g \delta_A^T \Sigma_{gAA}^{-1} \delta_A + C_2 (\max_g \delta_A^T \Sigma_{gAA}^{-1} \delta_A \vee \gamma) \left\{ s \log(\eta^{-1}) / \min(n_1, n_2) \right\}^{1/2} \right]^{-1}$$

$$\times \min_{j \in A} \left\| \pi_2 e_j^T S_{1AA}^{-1} d_A, \; \pi_1 e_j^T S_{2AA}^{-1} d_A \right\|_2.$$

Furthermore,

$$\min_{j \in A} \left\| \pi_2 e_j^T S_{1AA}^{-1} d_A, \; \pi_1 e_j^T S_{2AA}^{-1} d_A \right\|_2$$

$$= \min_{j \in A} \left\{ \pi_2^2 (e_j^T S_{1AA}^{-1} d_A)^2 + \pi_1^2 (e_j^T S_{2AA}^{-1} d_A)^2 \right\}^{1/2}$$

$$= \min_{j \in A} \left[ \pi_2^2 \{ e_j^T (S_{1AA}^{-1} d_A - \Sigma_{1AA}^{-1} \delta_A + \Sigma_{1AA}^{-1} \delta_A) \}^2 + \pi_1^2 \{ e_j^T (S_{2AA}^{-1} d_A - \Sigma_{2AA}^{-1} \delta_A + \Sigma_{2AA}^{-1} \delta_A) \}^2 \right]^{1/2}$$

$$\geq \min_{j \in A} \left\| \pi_2 e_j^T \Sigma_{1AA}^{-1} \delta_A, \; \pi_1 e_j^T \Sigma_{2AA}^{-1} \delta_A \right\|_2 - \max_g \left( \| S_{gAA}^{-1} d_A - \Sigma_{gAA}^{-1} \delta_A \|_\infty \right)$$

$$= \psi_{\min} - \max_g \left( \| S_{gAA}^{-1} d_A - \Sigma_{gAA}^{-1} \delta_A \|_\infty \right),$$

where in the last inequality we used $\pi_1^2 + \pi_2^2 \leq 1$. Using Lemma 8

$$\max_g \left( \| S_{gAA}^{-1} d_A - \Sigma_{gAA}^{-1} \delta_A \|_\infty \right)$$

$$\leq C \left[ \max_{j \in A, g} \left\{ (\Sigma_{gAA}^{-1})_{jj} (\delta_A^T \Sigma_{gAA}^{-1} \delta_A \vee \gamma) \right\} s \log(\eta^{-1}) / \min(n_1, n_2) \right]^{1/2}$$

with probability at least $1 - \eta$.

Therefore, to have $A \subseteq \widehat{A}$, it is sufficient to have

$$\psi_{\min} > C \left[ \max_{j \in A, g} \left\{ (\Sigma_{gAA}^{-1})_{jj} (\delta_A^T \Sigma_{gAA}^{-1} \delta_A \vee \gamma) \right\} s \log(\eta^{-1}) / \min(n_1, n_2) \right]^{1/2}$$

$$+ \left[ 1 + C_1 \max_g \delta_A^T \Sigma_{gAA}^{-1} \delta_A + C_2 (\max_g \delta_A^T \Sigma_{gAA}^{-1} \delta_A \vee \gamma) \left\{ s \log(\eta^{-1}) / \min(n_1, n_2) \right\} \right]$$

$$\times \lambda \max_g \| \Sigma_{gAA}^{-1} \|_2 s^{1/2} \left[ 1 + C \left\{ s \log(\eta^{-1}) / \min(n_1, n_2) \right\}^{1/2} \right].$$

Using the conditions on $\lambda$, and the fact that $\gamma \geq 1$, it follows that the second term above is the dominant term, and therefore it is sufficient to have for some constant $C > 0$

$$\psi_{\min} > C \lambda s^{1/2} \max_g \| \Sigma_{gAA}^{-1} \|_2 (\max_g \delta_A^T \Sigma_{gAA}^{-1} \delta_A \vee \gamma).$$

$\square$

# Appendix D  Supporting theorems and lemmas

**Theorem 5** (Oracle solution). *Consider an oracle estimator $[\tilde{v}_{1A}\ \tilde{v}_{2A}]$ from* (B.3). *Let $\rho_1 = n_1/n$, $\rho_2 = n_2/n$. Then*

$$\tilde{v}_{1A} = \rho_2 S_{1AA}^{-1} d_A (1 + \rho_2^2 d_A^{\mathrm{T}} S_{1AA}^{-1} d_A)^{-1} - \lambda \left(S_{1AA} + \rho_2^2 d_A d_A^{\mathrm{T}}\right)^{-1} \tilde{u}_{1A},$$

$$\tilde{v}_{2A} = \rho_1 S_{2AA}^{-1} d_A (1 + \rho_1^2 d_A^{\mathrm{T}} S_{2AA}^{-1} d_A)^{-1} - \lambda \left(S_{2AA} + \rho_1^2 d_A d_A^{\mathrm{T}}\right)^{-1} \tilde{u}_{2A};$$

*where $\tilde{u}_A$ is defined in* (B.4).

*Proof.* We present the proof only for $\tilde{v}_{1A}$, the proof for $\tilde{v}_{2A}$ is analogous. From Section 2.3

$$[\tilde{v}_{1A}\ \tilde{v}_{2A}] = \underset{v_{1A}, v_{2A} \in \mathbb{R}^s}{\mathrm{argmin}} \left\{ \widehat{L}_{\psi_1}(v_{1A}) + \widehat{L}_{\psi_2}(v_{2A}) + \lambda \sum_{j=1}^s (v_{1Aj}^2 + v_{2Aj}^2)^{1/2} \right\},$$

$$\widehat{L}_{\psi_1}(v_{1A}) + \widehat{L}_{\psi_2}(v_{2A})$$
$$= v_{1A}^{\mathrm{T}} S_{1AA} v_{1A}/2 + (n_2/n d_A^{\mathrm{T}} v_{1A} - 1)^2/2 + v_{2A}^{\mathrm{T}} S_{2AA} v_{2A}/2 + (n_2/n d_A^{\mathrm{T}} v_{2A} - 1)^2/2.$$

Using the optimality conditions, the oracle solution must satisfy

$$\tilde{v}_{1A} = \left(S_{1AA} + \rho_2^2 d_A d_A^{\mathrm{T}}\right)^{-1} (\rho_2 d_A - \lambda \tilde{u}_{1A}),$$

where $\tilde{u}_A$ is the subgradient of $\sum_{j=1}^s (v_{1Aj}^2 + v_{2Aj}^2)^{1/2}$ in (B.4). By Sherman–Morrison identity,

$$(S_{1AA} - \rho_2^2 d_A d_A^{\mathrm{T}})^{-1} = S_{1AA}^{-1} - (1 + \rho_2^2 d_A^{\mathrm{T}} S_{1AA}^{-1} d_A)^{-1} \rho_2^2 S_{1AA}^{-1} d_A d_A^{\mathrm{T}} S_{1AA}^{-1}.$$

The statement follows by combining the above two displays. □

**Lemma 1.** *There exist constant $C > 0$ such that with probability at least $1 - \eta$*

$$|n_g/n - \pi_g| \le C \left\{ \log(\eta^{-1})/n \right\}^{1/2} \quad (g = 1,2), \quad |n_1/n_2 - \pi_1/\pi_2| \le C \left\{ \log(\eta^{-1})/n \right\}^{1/2}.$$

*Proof.* Given that $n_g \sim \mathrm{Bin}(n, \pi_g)$, by Hoeffding inequality $\mathrm{pr}(|\pi_g - n_g/n| \ge \varepsilon) \le 2 \exp(-2n\varepsilon^2)$. Let $\eta = 2\exp(-2n\varepsilon^2)$, then $2n\varepsilon^2 = \log(2\eta^{-1})$, $\varepsilon = C\{\log(\eta^{-1})/n\}^{1/2}$ and $n_g/n = \pi_g + \mathcal{O}_p\{\log(\eta^{-1})/n\}^{1/2}$. Let $f(x) = x/(1-x)$, which is non-decreasing for $x \in (0,1)$. Since $n_1/n_2 = f(n_1/n)$, the second inequality in the lemma follows from the first. □

**Lemma 2.** *Let $E_{Ug} = U_{gA^c} - \Sigma_{gA^cA}\Sigma_{gAA}^{-1}U_{gA}$, $E_{dg} = d_{A^c} - \Sigma_{gA^cA}\Sigma_{gAA}^{-1}d_A$, $g = 1,2$. Then $E_{Ug}$ is independent from $U_{gA}$, $E_{Ug} \sim \mathcal{N}(0, \Sigma_{gA^cA^c:A} \otimes I_{n_g-1})$, $e_j^{\mathrm{T}} E_{dg} \sim \mathcal{N}\left(0, n_g^{-1} \sigma_{jdg}^2\right)$; where $\sigma_{jdg}^2 = e_j^{\mathrm{T}} \Sigma_{d_g} e_j$, and $\Sigma_{gA^cA^c:A}$, $\Sigma_{d_g}$ are defined in* (B.2).

*Proof.* Since $E_{dg}$, $E_{Ug}$ are formed by applying linear transformation to normal $d$, $U_1$, $U_2$, it follows that $E_{dg}$, $E_{Ug}$ are also normally distributed. It remains to verify the form of the means and covariance matrices. We consider $g = 1$, the proof for $g = 2$ is similar.

Consider $E_{U1}$. By definition, the columns of $U_1$ satisfy $u_{1i} \sim N(0, \Sigma_1)$. Since

$$E_{U1} = (-\Sigma_{1A^cA}\Sigma_{1AA}^{-1}\ I_{p-s}) \begin{pmatrix} U_{1A} \\ U_{1A^c} \end{pmatrix},$$

it follows that $\mathbb{E}(E_{U1}) = 0$, and

$$\operatorname{var}(E_{U1}) = (-\Sigma_{1A^cA}\Sigma_{1AA}^{-1}\ I_{p-s}) \begin{pmatrix} \Sigma_{1AA} & \Sigma_{1AA^c} \\ \Sigma_{1A^cA} & \Sigma_{1A^cA^c} \end{pmatrix} (-\Sigma_{1A^cA}\Sigma_{1AA}^{-1}\ I_{p-s})^{\mathrm{T}} \otimes I_{n_1-1}$$
$$= (\Sigma_{1A^cA^c} - \Sigma_{1A^cA}\Sigma_{1AA}^{-1}\Sigma_{1AA^c}) \otimes I_{n_1-1}.$$

Consider $E_{d1}$. Since $\Sigma_1^{-1}\delta = \psi_1 = (\psi_{1A}^{\mathrm{T}}, 0)^{\mathrm{T}}$, by rewriting $\Sigma_1\Sigma_1^{-1}\delta = \delta$, and using block matrices of $\Sigma_1$ and $\Sigma_1^{-1}$, it follows that $\Sigma_{1A^cA}\Sigma_{1AA}^{-1}\delta_A = \delta_{A^c}$. Then $\mathbb{E}(E_{d1}) = \delta_{A^c} - \Sigma_{1A^cA}\Sigma_{1AA}^{-1}\delta_A = 0$. Furthermore,

$$\operatorname{var}(E_{d1})$$
$$= \operatorname{var}(d_{A^c} - \Sigma_{1A^cA}\Sigma_{1,AA}^{-1}d_A)$$
$$= \operatorname{var}(d_{A^c}) + \Sigma_{1A^cA}\Sigma_{1AA}^{-1}\operatorname{var}(d_A)\Sigma_{1AA}^{-1}\Sigma_{1AA^c}$$
$$\quad - \Sigma_{1A^cA}\Sigma_{1AA}^{-1}\operatorname{cov}(d_A, d_{A^c}) - \operatorname{cov}(d_{A^c}, d_A)\Sigma_{1AA}^{-1}\Sigma_{1AA^c}$$
$$= n_1^{-1}\Sigma_{1A^cA^c} + n_2^{-1}\Sigma_{2A^cA^c} + \Sigma_{1A^cA}\Sigma_{1AA}^{-1}\left(n_1^{-1}\Sigma_{1AA} + n_2^{-1}\Sigma_{2AA}\right)\Sigma_{1AA}^{-1}\Sigma_{1AA^c}$$
$$\quad - \Sigma_{1A^cA}\Sigma_{1AA}^{-1}\left(n_1^{-1}\Sigma_{1AA^c} + n_2^{-1}\Sigma_{2AA^c}\right) - \left(n_1^{-1}\Sigma_{1A^cA} + n_2^{-1}\Sigma_{2A^cA}\right)\Sigma_{1AA}^{-1}\Sigma_{1AA^c}$$
$$= n_1^{-1}\Sigma_{1A^cA^c:A} + n_2^{-1}\Big(\Sigma_{2A^cA^c} + \Sigma_{1A^cA}\Sigma_{1AA}^{-1}\Sigma_{2AA}\Sigma_{1AA}^{-1}\Sigma_{1AA^c}$$
$$\quad\quad - \Sigma_{1A^cA}\Sigma_{1AA}^{-1}\Sigma_{2AA^c} - \Sigma_{2A^cA}\Sigma_{1AA}^{-1}\Sigma_{1AA^c}\Big).$$

□

**Lemma 3.** *Let $S_{gAA}$ be a submatrix of the sample covariance matrix for group $g \in \{1, 2\}$ corresponding to variables in $A$, with $s = \operatorname{card}(A)$. Let $\Sigma_{gAA}$ be the corresponding submatrix of population covariance matrix. Under Assumption 1, there exist constants $C_1, C_2 > 0$ such that with probability at least $1 - \eta$*

$$\|\Sigma_{gAA}^{1/2}S_{gAA}^{-1}\Sigma_{gAA}^{1/2} - I\|_2 \leq C_1\left\{s\log(\eta^{-1})/n_g\right\}^{1/2}, \quad \|S_{gAA}^{-1}\|_2 \leq \|\Sigma_{gAA}^{-1}\|_2\left[1 + C_2\left\{s\log(\eta^{-1})/n_g\right\}^{1/2}\right].$$

*Proof.* Using normality, the sample covariance matrices satisfy $S_{gAA} = (n_g - 1)^{-1}W_g W_g^{\mathrm{T}}$ with $W_g \in \mathbb{R}^{s \times (n_g - 1)}$ having independent columns $w_{gi} \sim N(0, \Sigma_{gAA})$. Then the desired bounds follow from Wainwright (2009, Lemma 9). □

**Lemma 4.** *Let a random vector $X \in \mathbb{R}^s$ be such that $X \sim \mathcal{N}(0, n^{-1}A)$. Then there exist constant $C > 0$ such that with probability at least $1 - \eta$*

$$\|X\|_2 \leq C\left\{\|A\|_2 n^{-1}s\log(\eta^{-1})\right\}^{1/2}.$$

*Proof.* Since $A^{-1/2}X \sim \mathcal{N}(0, n^{-1}I_s)$, by Hsu et al. (2012, Proposition 1.1), with probability at least $1 - \eta$

$$\|A^{-1/2}X\|_2^2 \leq s/n + 2\left\{s\log(\eta^{-1})\right\}^{1/2}/n + 2\log(\eta^{-1})/n.$$

For small $\eta$ it follows that there exist $C > 0$ such that $\|A^{-1/2}X\|_2^2 \leq Cn^{-1}s\log(\eta^{-1})$ with probability at least $1 - \eta$. The statement of the lemma follows since

$$\|X\|_2^2 = X^{\mathrm{T}}X = X^{\mathrm{T}}A^{-1/2}AA^{-1/2}X \leq \|A\|_2\|A^{-1/2}X\|_2^2.$$

□

**Lemma 5.** *There exist constant $C > 0$ such that with probability at least $1 - \eta$*

$$\max_g \|\Sigma_{gAA}^{-1/2}(d_A - \delta_A)\|_2 \leq C\left\{\gamma s \log(\eta^{-1})/\min(n_1, n_2)\right\}^{1/2},$$

*where $\gamma$ is defined in* (B.1).

*Proof.* Since $d_A - \delta_A \sim \mathcal{N}(0, n_1^{-1}\Sigma_{1AA} + n_2^{-1}\Sigma_{2AA})$, it follows that

$$\Sigma_{1AA}^{-1/2}(d_A - \delta_A) \sim \mathcal{N}\left(0, n_1^{-1}\left(I + n_2^{-1}n_1\Sigma_{1AA}^{-1/2}\Sigma_{2AA}\Sigma_{1AA}^{-1/2}\right)\right).$$

Applying Lemma 1 and Lemma 4 concludes the proof. The case $g = 2$ is analogous. $\square$

**Lemma 6.** *There exist constants $C_1, C_2$ such that with probability at least $1 - \eta$ for $g = 1, 2$*

$$d_A^{\mathrm{T}} S_{gAA}^{-1} d_A \leq C_1 d_A^{\mathrm{T}} \Sigma_{gAA}^{-1} d_A \left[1 + C_2\left\{\log(\eta^{-1})/(n_g - s)\right\}^{1/2}\right].$$

*Proof.* We prove for $g = 1$, case $g = 2$ is analogous. Since $(n_1 - 1)S_{1AA} \sim W_s(n_1 - 1, \Sigma_{1AA})$, and $d_A$ is independent of $S_{1AA}$, by Muirhead (1982, Theorem 3.2.12)

$$(n_1 - 1)\frac{d_A^{\mathrm{T}}\Sigma_{1AA}^{-1}d_A}{d_A^{\mathrm{T}}S_{1AA}^{-1}d_A} \sim \chi_{n_1-s}^2.$$

Using (Laurent and Massart, 2000, Lemma 1),

$$\mathrm{pr}\left[(n_1 - 1)\frac{d_A^{\mathrm{T}}\Sigma_{1AA}^{-1}d_A}{d_A^{\mathrm{T}}S_{1AA}^{-1}d_A} \geq (n_1 - s) - 2\left\{(n_1 - s)\log(\eta^{-1})\right\}^{1/2}\right] \geq 1 - \eta.$$

Therefore, with probability at least $1 - \eta$

$$d_A^{\mathrm{T}} S_{1AA}^{-1} d_A \leq (n_1 - 1)(n_1 - s)^{-1} d_A^{\mathrm{T}} \Sigma_{1AA}^{-1} d_A \left[1 - 2\left\{\log(\eta^{-1})/(n_1 - s)\right\}^{1/2}\right]^{-1}.$$

Hence, there exist constants $C_1, C_2 > 0$ such that with probability at least $1 - \eta$

$$d_A^{\mathrm{T}} S_{1AA}^{-1} d_A \leq C_1 d_A^{\mathrm{T}} \Sigma_{1AA}^{-1} d_A \left[1 + C_2\left\{\log(\eta^{-1})/(n_1 - s)\right\}^{1/2}\right].$$

$\square$

**Lemma 7.** *There exist constant $C > 0$ such that with probability at least $1 - \eta$*

$$d_A^{\mathrm{T}}\Sigma_{gAA}^{-1}d_A \leq C\left\{\delta_A^{\mathrm{T}}\Sigma_{gAA}^{-1}\delta_A + \gamma n_g^{-1} s \log(\eta^{-1})\right\} \quad (g = 1, 2),$$

*where $\gamma$ is defined in* (B.1).

*Proof.* We prove the result for $g = 1$, the case $g = 2$ is similar. Consider

$$d_A^{\mathrm{T}}\Sigma_{1AA}^{-1}d_A = \delta_A^{\mathrm{T}}\Sigma_{1AA}^{-1}\delta_A + 2(d_A - \delta_A)^{\mathrm{T}}\Sigma_{1AA}^{-1}\delta_A + (d_A - \delta_A)^{\mathrm{T}}\Sigma_{1AA}^{-1}(d_A - \delta_A)$$
$$\leq 2\delta_A^{\mathrm{T}}\Sigma_{1AA}^{-1}\delta_A + 2(d_A - \delta_A)^{\mathrm{T}}\Sigma_{1AA}^{-1}(d_A - \delta_A).$$

By Lemma 5, there exist constant $C \geq 0$ such that with probability at least $1 - \eta$

$$(d_A - \delta_A)^{\mathrm{T}}\Sigma_{1AA}^{-1}(d_A - \delta_A) \leq C\gamma n_1^{-1} s \log(\eta^{-1}).$$

The result follows by combining the above displays. $\square$

**Corollary 1.** *There exist constants $C_1, C_2, C_3 > 0$ such that with probability at least $1 - \eta$ for $g = 1, 2$ and $\gamma$ in (B.1)*

$$d_A^T S_{gAA}^{-1} d_A \leq C_1 \delta_A^T \Sigma_{gAA}^{-1} \delta_A \left[1 + C_2 \left\{\log(\eta^{-1})/(n_g - s)\right\}^{1/2}\right] + C_3 \gamma n_g^{-1} s \log(\eta^{-1}).$$

*Proof.* The result follows by combining results of Lemma 6 and Lemma 7. □

**Lemma 8.** *There exist constant $C > 0$ such that with probability at least $1 - \eta$ for $g = 1, 2$*

$$\|S_{gAA}^{-1} d_A - \Sigma_{gAA}^{-1} \delta_A\|_\infty \leq C \left\{\max_{j \in A}(\Sigma_{gAA}^{-1})_{jj}(\delta_A^T \Sigma_{gAA}^{-1} \delta_A \vee \gamma) n_g^{-1} s \log(\eta^{-1})\right\}^{1/2},$$

*where $\gamma$ is defined in* (B.1).

*Proof.* We prove the result for $g = 1$, the case $g = 2$ is similar. Consider

$$\begin{aligned}
&|e_j^T S_{1AA}^{-1} d_A - e_j^T \Sigma_{1AA}^{-1} \delta_A| \\
&= |e_j^T (S_{1AA}^{-1} - \Sigma_{1AA}^{-1})(d_A - \delta_A) + e_j^T (S_{1AA}^{-1} - \Sigma_{1AA}^{-1})\delta_A + e_j^T \Sigma_{1AA}^{-1}(d_A - \delta_A)| \\
&\leq (e_j^T \Sigma_{1AA}^{-1} e_j)^{1/2} \|(\Sigma_{1AA}^{1/2} S_{1AA}^{-1} \Sigma_{1AA}^{1/2} - I)\Sigma_{1AA}^{-1/2}(d_A - \delta_A)\|_2 \\
&\quad + (e_j^T \Sigma_{1AA}^{-1} e_j)^{1/2} \|(\Sigma_{1AA}^{1/2} S_{1AA}^{-1} \Sigma_{1AA}^{1/2} - I)\Sigma_{1AA}^{-1/2} \delta_A\|_2 \\
&\quad + (e_j^T \Sigma_{1AA}^{-1} e_j)^{1/2} \|\Sigma_{1AA}^{-1/2}(d_A - \delta_A)\|_2.
\end{aligned}$$

Let $m_1 = \|\Sigma_{1AA}^{1/2} S_{1AA}^{-1} \Sigma_{1AA}^{1/2} - I\|_2$ and $m_2 = \|\Sigma_{1AA}^{-1/2}(d_A - \delta_A)\|_2$. Using the above display

$$\|S_{1AA}^{-1} d_A - \Sigma_{1AA}^{-1} \delta_A\|_\infty \leq \max_{j \in A}(\Sigma_{1AA}^{-1})_{jj}^{1/2} \left\{m_1 m_2 + m_1 (\delta_A^T \Sigma_{1AA}^{-1} \delta_A)^{1/2} + m_2\right\}. \quad (C.1)$$

Using Lemma 3, there exist constant $C_1 > 0$ such that $m_1 \leq C_1 \{s \log(\eta^{-1})/n_1\}^{1/2}$ with probability at least $1 - \eta$. Using Lemma 5, there exist constant $C_2 > 0$ such that $m_2 \leq C_2 \{\gamma s \log(\eta^{-1})/n_1\}^{1/2}$ with probability at least $1 - \eta$. Combining these bounds with (C.1), there exist constant $C > 0$ such that with probability at least $1 - \eta$

$$\|S_{1AA}^{-1} d_A - \Sigma_{1AA}^{-1} \delta_A\|_\infty \leq C \left\{\max_{j \in A}(\Sigma_{1AA}^{-1})_{jj}(\delta_A^T \Sigma_{1AA}^{-1} \delta_A \vee \gamma) n_1^{-1} s \log(\eta^{-1})\right\}^{1/2}.$$

□